\documentclass[10pt,twocolumn,letterpaper]{article}

\usepackage[pagenumbers]{cvpr} 
\usepackage[accsupp]{axessibility}
\usepackage{graphicx}
\usepackage{amsmath}
\usepackage{amssymb}
\usepackage{booktabs}

\usepackage[pagebackref,breaklinks,colorlinks]{hyperref}

\usepackage[capitalize]{cleveref}
\crefname{section}{Sec.}{Secs.}
\Crefname{section}{Section}{Sections}
\Crefname{table}{Table}{Tables}
\crefname{table}{Tab.}{Tabs.}

\usepackage{times}
\usepackage{amsmath}
\usepackage{hyperref}
\usepackage{algorithmic}
\usepackage{algorithm}
\usepackage{amssymb}
\usepackage{graphicx}
\usepackage{epsfig}
\usepackage{wrapfig,lipsum,booktabs}
\usepackage{footnote}
\usepackage{enumitem}
\usepackage{array}
\usepackage{changepage}
\usepackage{mathtools}

\newtheorem{theorem}{Theorem}
\newtheorem{lemma}{Lemma}
\newtheorem{proposition}{Proposition}

\begin{document}

\title{Hierarchical Correlation Clustering and Tree Preserving Embedding}

\author{Morteza Haghir Chehreghani\\
Chalmers University of Technology\\
Gothenburg, Sweden\\
{\tt\small morteza.chehreghani@chalmers.se}
\and
Mostafa Haghir Chehreghani\\
Amirkabir University of Technology (Tehran Polytechnic)\\
Tehran, Iran\\
{\tt\small mostafa.chehreghani@gmail.com}
}
\maketitle

\begin{abstract}
We propose a hierarchical correlation clustering method that extends the well-known correlation clustering to produce hierarchical clusters applicable to both positive and negative pairwise dissimilarities. Then, in the following, we study unsupervised representation learning with such hierarchical correlation clustering. For this purpose, we first investigate embedding the respective hierarchy to be used for tree preserving embedding and feature extraction. Thereafter, we study the extension of minimax distance measures to correlation clustering, as another representation learning paradigm. Finally, we demonstrate the performance of our methods on several datasets.
\end{abstract}

\section{Introduction}
Data clustering plays an essential role in unsupervised learning and exploratory data analysis. It is used in a variety of applications including web mining, network analysis, image segmentation, bioinformatics, user analytics and knowledge management. Its goal is to partition the data into groups in a way that the objects in the same cluster are more similar according to some criterion, compared to the objects in different clusters.

Many  clustering methods partition the data into $K$ flat clusters 
for example,  $K$-means \cite{mcqueen1967smc}, spectral clustering \cite{Shi:2000:NCI,Ng01onspectral} 
and correlation clustering \cite{BansalBC04}. In many applications, however, the clusters are preferred to be presented at different levels, encompassing both high-level and detailed information.  Hierarchical clustering is useful to produce such structures, usually encoded by a \emph{dendrogram}. 
A dendrogram is a tree data structure where each node corresponds to a cluster, with the leaf nodes (those at the bottom of the tree) containing only one object. Higher-level clusters are formed by aggregating lower-level clusters and the inter-cluster dissimilarity between them.

Hierarchical clustering can be performed either in an agglomerative (i.e., bottom-up) or in a divisive (i.e., top-down) manner \cite{Maimon:2005}.
Agglomerative methods  are often computationally more efficient, making them more popular in practice \cite{podani2000introduction}.
In both approaches, the clusters are aggregated or split based on various criteria, such as \emph{single}, \emph{average}, \emph{centroid}, \emph{complete} and \emph{Ward}.
Several  studies aim to improve these methods. 
The works in \cite{biom12647,Levenstien2003} focus on the statistical significance of hierarchical clustering. 
\cite{Cohen-Addad:2018:HCO,NIPS2017_7200,Roy:2017:HCV} formulate this problem as an optimization problem and propose  approximate solutions.  \cite{Yildirim2017} considers multiple dissimilarities for a pair of clusters, and \cite{Bruynooghe1977,ChehreghaniIJCNN21} suggest merging multiple clusters at each step instead of one.  \cite{Balcan:2014:RHC} employs global information to  eliminate the influence of noisy similarities, 
and \cite{ChehreghaniAC08} proposes to apply agglomerative methods to small subsets of the data instead of individual data objects.
\cite{doi:10.1198,jcgs2012} augment agglomerative methods with probabilistic models, and finally, \cite{abs-1109-2378,Cochez:2015:TTA} propose efficient but approximate methods for hierarchical clustering.

On the other hand, most clustering methods, either flat or hierarchical, assume non-negative pairwise (dis)similarities.  However, in several practical applications, pairwise similarities can be any real number, positive or negative. For example, it could be preferable for a user or oracle to indicate whether two objects are similar (considered a positive relation) or dissimilar (considered a negative relation), rather than solely providing a positive (non-negative) pairwise similarity, even if the two objects are dissimilar.
The former approach yields more precise information because, in the latter scenario, the dissimilarity between two objects (i.e., zero similarity) could be confused with a lack of available information.  
Some relevant applications for this setting include image segmentation with higher order correlation information \cite{KimNKY11,KimYNK14}, webpage segmentation \cite{ChakrabartiKP08}, community detection over graphs \cite{ShiDELM21},  social media mining \cite{TangCAL16}, analysis of connections over web \cite{KalashnikovCMN08}, dealing with attraction/rejection data \cite{DemaineEFI06}, automated label generation from clicks \cite{AgrawalHKMT09} and entity resolution \cite{BansalBC02,2367502.2367564}.

Hence, a specialized clustering model known as \emph{correlation clustering} has been developed to work with such data. This model was first introduced on the graphs with only $+1$ or $-1$ pairwise similarities \cite{BansalBC04,BansalBC02}, and then was generalized to the graphs with arbitrary positive or negative edge weights \cite{DemaineEFI06,AilonCN08,CGW03}.
The original model obtains the number of clusters automatically. The variant in \cite{ChehreghaniAISTATS12} limits the number of clusters to fixed $K$ clusters. Semidefinite programming (SDP) relaxation provides tight approximation bounds in particular for maximizing the agreements \cite{CGW03,MS10}, although it is computationally inefficient in practice \cite{ThielCD19}.
Then, \cite{ChehreghaniICDM17,ThielCD19} provide efficient greedy algorithms based on local search and Frank-Wolfe optimization with a fast convergence rate.

However, all of these methods produce flat correlation clusters.
In this paper, 
we first propose a \emph{Hierarchical Correlation Clustering} (HCC) method that handles both positive and negative  pairwise (dis)similarities and produces clusters at different levels  (Section 3).  To the best of our knowledge, this work is one the first extensions of the well-known correlation clustering  to  hierarchical clustering.\footnote{We note the so-called hierarchical correlation clustering methods proposed in \cite{AchtertBKZ06,GuW10,LiebmannWS18} are irrelevant to the well-studied correlation clustering problem \cite{BansalBC04,BansalBC02}; they  study for example the correlation coefficients for high-dimensional data.} 
A hierarchical correlation clustering, also called HCC, is developed in \cite{VainsteinCCRMA21}. This method offers a $0.4767$ approximation, but lacking experimental evaluation. Furthermore,  unlike our method, the method in \cite{VainsteinCCRMA21} does not follow the generic agglomerative clustering procedure. 

We note that unlike flat clustering, hierarchical clustering yields an ordering among the objects, i.e., objects that join earlier in the hierarchy are closer to each other than those that join at later steps.  
This implies that hierarchical clustering induces a new (dis)similarity measure between the objects, connected to the way the objects join each other to form clusters at different levels.  
Thereby, in the following, we consider two representation learning methods related to hierarchical clustering and study their adaptation to hierarchical correlation clustering. This enables us to not only use HCC for producing hierarchical clusters, but also to employ it for computing a suitable similarity/distance measure as an intermediate data processing step.

One way to perform representation learning from hierarchical clustering is to compute an embedding that corresponds to the respective hierarchy.  \emph{Tree preserving embedding} \cite{ShiehHA11,Shieh16916} is a method that achieves this for the special case of single linkage method. 
Later, \cite{ChehreghaniMLJ20} develops tree preserving embedding for various standard agglomerative clustering methods.
We then adapt these works (in particular the later work \cite{ChehreghaniMLJ20}) to develop a tree preserving embedding for HCC dendrograms (Section 4) where the embedded features can be used as a set of new features for an arbitrary downstream task.
 In this way, we can 
investigate HCC for the purpose of computing relevant features for a probabilistic method such as Gaussian Mixture Model  (GMM), instead of solely using HCC for the purpose of hierarchical clustering.
This enables us to apply a method like GMM for clustering the pairwise similarities that can be positive or negative numbers, a task that was not possible before.

Another representation learning paradigm that we study is called \emph{minimax} dissimilarity, a graph-based method that is tightly connected to hierarchical clustering. 
It provides a sophisticated way to infer transitive relations and extract manifolds and elongated clusters in an unsupervised way \cite{FischerB03,KimC07icml,Chehreghani17AAAI,LittleMM20}. 
Thereby, for the first time, we study minimax dissimilarities on the graphs with positive and negative (dis)similarities, i.e., with correlation clustering (Section 5).  We show that using minimax dissimilarities with correlation clustering not only helps for extracting elongated patterns, but also yields a significant reduction in the computational complexity, i.e., from NP-hardness to a polynomial runtime.

We finally perform several experiments on various datasets to demonstrate the effectiveness of our methods in different settings (Section 6).

\section{Notations and Definitions}

A dataset is characterized by a set of $n$ objects with indices $\mathbf O= \{1,...,n\}$ and a pairwise similarity or dissimilarity matrix.  An $n \times n$ matrix $\mathbf S \in \mathbb R^{n\times n}$ represents the pairwise similarities between the objects, whereas, the pairwise dissimilarities are shown by matrix $\mathbf D \in \mathbb R^{n\times n}$. Both of similarities and dissimilarities can be positive or negative. This property allows us to convert the pairwise similarities to dissimilarities by a simple transformation such as  $\mathbf D = -\mathbf S$, i.e., the pairwise dissimilarities are obtained by the negation of the  similarities and vice versa. 
\footnote{Such a \emph{nonparametric} transformation resolves the issues related to obtaining a proper similarity measure from pairwise dissimilarities. For example, with kernels, e.g., RBF kernels, finding the optimal parameter(s) is often crucial and nontrivial, and the optimal parameters occur inside a very narrow range \cite{Nadler07,Luxburg07}. Moreover, the methods we develop in this paper are unaffected by the choice of the transformation $\mathbf D$; for example in Algorithm \ref{alg:hierarchical_CC}, we only use the pairwise similarities $\mathbf{S}$.}
The objects and the pairwise (dis)similarities are represented by graph $\mathcal{G} = (\mathbf O, \mathbf  S)$ or $\mathcal{G} = (\mathbf O, \mathbf  D)$.

A cluster is represented by a set, e.g.,  $\mathbf v$, which includes the objects belong to that. The function $dis(\mathbf u,\mathbf v)$ denotes the inter-cluster dissimilarity between clusters $\mathbf u$ and $\mathbf v$  that can be defined according to different criteria.
A hierarchical  clustering  solution can be represented by a dendrogram $T$ defined as a rooted ordered tree such that, 
i) each node $\mathbf v$ in $T$ includes a non-empty subset of the objects corresponding to a cluster, i.e., $\mathbf v \subseteq \mathbf O,|\mathbf v| \ge 1, \forall \mathbf v \in T$, with the leaf nodes including distinct single objects, and
ii) the overlapping clusters are ordered, i.e., $\forall \mathbf u,\mathbf v \in T, \text{if } \mathbf u \cap \mathbf v \ne 0, \text{ then either } \mathbf u \subseteq \mathbf v \text{ or } \mathbf v \subseteq \mathbf u$. 
The latter condition implies that between every two overlapping nodes an ancestor-descendant relation holds, i.e., $\mathbf u \subseteq \mathbf v$ indicates $\mathbf v$ is an ancestor of $\mathbf u$, and $\mathbf u$ is a descendant of $\mathbf v$.

The clusters at the lowest level, called \emph{leaf} clusters/node, are the individual distinct objects, i.e., $\mathbf v$ is a leaf cluster if and only if $|\mathbf v|=1$.  A cluster at a higher level contains the union of the objects of its children. The root of a dendrogram is defined as the cluster at the highest level which has the maximum size, i.e., all other clusters are its descendants. $linkage(\mathbf v), \mathbf v \in T$ returns the dissimilarity between the children of $\mathbf v$ based on the criterion used to compute the dendrogram (i.e., $dis(c_l,c_r)$ where $c_l$ and $c_r$ indicate the two child clusters of $\mathbf v$).
For simplicity of explanation, w.l.g., we assume every non-leaf cluster has two children. The level of cluster $\mathbf v$, i.e., $level(\mathbf v)$, is determined by

\begin{equation}
level(\mathbf v) = \max(level(c_l), level(c_r))+1.
\end{equation}

For the leaf clusters,  $level()$ and $dis()$  return $0$.
Every connected subtree of $T$
whose leaf clusters contain only individual objects from $\mathbf O$ constitutes a dendrogram on this subset of objects. We require that every common node present in both $T$ and the subtree must have the same child nodes or clusters. We use $\mathcal T_T$ to refer to the set of all (sub)dendrograms obtained in this way from $T$.

\section{Hierarchical Correlation Clustering}
Agglomerative methods begin with each object in a separate cluster, and then at each round, combine the two clusters that have a \emph{minimal} dissimilarity according to a criterion (defined by the $dis(.,.)$ function) until only one cluster remains. 
For example, the \emph{single} linkage (SL) criterion \cite{sneath1957dn09j} defines the dissimilarity between two clusters as the dissimilarity between their nearest members ($dis(\mathbf u,\mathbf v) = \min_{i \in \mathbf u, j \in \mathbf v} \mathbf D_{i,j}$), whereas,  \emph{complete} linkage (CL)  \cite{lance67hierarchical} uses the dissimilarity between their farthest members ($dis(\mathbf u,\mathbf v) = \max_{i \in \mathbf u, j \in \mathbf v} \mathbf D_{i,j}$). On the other hand, the \emph{average} linkage (AL)  criterion \cite{sokal58} considers the average of the inter-cluster dissimilarities as the dissimilarity between the two clusters ($dis(\mathbf u,\mathbf v) = \sum_{i \in \mathbf u, j \in \mathbf v} \frac{\mathbf D_{i,j}}{|\mathbf u||\mathbf v|}$).
These methods can be shown to be shift-invariant, as mentioned in  Proposition \ref{thm:agg_shift} \cite{MortezaShift_MLj}.

\begin{proposition}
  Single linkage, complete linkage and average linkage methods are invariant w.r.t. the shift of the pairwise dissimilarities by an arbitrary real number $\alpha$.
  \label{thm:agg_shift}
\end{proposition}

Thus, we can still use these methods even with possibly negative pairwise dissimilarities as shifting the pairwise dissimilarities (by a large enough constant) to make them non-negative does not change the solution.

However, clustering the data consisting of positive and negative dissimilarities is usually conducted by \emph{correlation clustering}. Thus, despite the applicability of single linkage, average linkage and complete linkage methods, we propose a novel hierarchical clustering consistent with the standard correlation clustering, called Hierarchical Correlation Clustering (HCC). This method is thus adapted to positive/negative pairwise (dis)similarities, and as our experiments confirm, it outperforms the other methods (i.e., SL, CL, and AL) when applied to such data.

The cost function for flat (standard) correlation clustering accounts for disagreements  (i.e., negative similarities inside clusters and positive similarities between clusters) and is written by \cite{ChehreghaniAISTATS12}
\begin{eqnarray}
&& R^{CC}(\mathbf v_1, ... , \mathbf v_K ;\mathbf S) =  \frac{1}{2}\sum_{k=1}^K \sum_{i,j \in \mathbf v_k} (|\mathbf S_{ij}|-\mathbf S_{ij})   \nonumber  \\
&&\qquad\qquad
 + \frac{1}{2}\sum_{k=1}^{K} \sum_{\substack{k'=1,\\k'\ne k}}^{K} \sum_{i\in\mathbf v_k} \sum_{j \in\mathbf v_{k'}} (|\mathbf S_{ij}|+ \mathbf S_{ij}),
\label{eq:fixedCC}
\end{eqnarray}
where $K$ is the number of clusters and $\mathbf v_k$'s indicate the different clusters.

We may rewrite the cost function as

\begin{align}
& R^{CC}(\mathbf v_1, ... , \mathbf v_K ;\mathbf S) =  -  \underbrace{\frac{1}{2}\sum_{k=1}^{K} \sum_{k'=1}^{K} \sum_{i\in\mathbf v_k} \sum_{j \in\mathbf v_{k'}}  \mathbf S_{ij} }_{constant} \nonumber \\
& \; + \underbrace{\frac{1}{2}\sum_{k=1}^K \sum_{i,j \in \mathbf v_k} |\mathbf S_{ij}| +  \frac{1}{2}\sum_{k=1}^{K} \sum_{\substack{k'=1,\\k'\ne k}}^{K} \sum_{i\in\mathbf v_k} \sum_{j \in\mathbf v_{k'}}|\mathbf S_{ij}|}_{constant}  \nonumber \\
 &  +  \frac{1}{2}\sum_{k=1}^{K} \sum_{\substack{k'=1,\\k'\ne k}}^{K} \sum_{i\in\mathbf v_k} \sum_{j \in\mathbf v_{k'}} \mathbf S_{ij} +  \frac{1}{2}\sum_{k=1}^{K} \sum_{\substack{k'=1,\\k'\ne k}}^{K} \sum_{i\in\mathbf v_k} \sum_{j \in\mathbf v_{k'}} \mathbf S_{ij}.
\end{align}

We then have

\begin{align}
&R^{CC}(\mathbf v_1, ... , \mathbf v_K ;\mathbf S) =
 constant + \sum_{k=1}^{K} \sum_{\substack{k'=1,\\k'\ne k}}^{K} \sum_{i\in\mathbf v_k} \sum_{j \in\mathbf v_{k'}} \mathbf S_{ij} \nonumber \\
 &\qquad\qquad\equiv
constant - \sum_{k=1}^{K} \sum_{\substack{k'=1,\\k'\ne k}}^{K} \sum_{i\in\mathbf v_k} \sum_{j \in\mathbf v_{k'}} \mathbf D_{ij} \,.
\label{eq:MinCut}
\end{align}

Therefore, correlation clustering aims to minimize the inter-cluster similarities, and in other words, it maximizes the inter-cluster dissimilarities.
This formulation in Eq. \ref{eq:MinCut}  inspires us for a  \emph{consistent} way of defining a new inter-cluster dissimilarity function for hierarchical (agglomerative) correlation clustering of positive and negative (dis)similarities.  At each step, we  merge the two clusters that have a minimal dissimilarity (or a maximal similarity), where we define the dissimilarity between the two clusters  $\mathbf u$ and $\mathbf v$ as
\begin{equation}
dis^{CC}(\mathbf u,\mathbf v) = \sum_{i\in\mathbf u} \sum_{j \in\mathbf v} \mathbf D_{ij}  = - \sum_{i\in\mathbf u} \sum_{j \in\mathbf v} \mathbf S_{ij}.
\label{eq:distCC}
\end{equation}

We emphasize that HCC is consistent with the generic agglomerative clustering framework applied with, for example, \emph{single} linkage, \emph{average} linkage, \emph{complete} linkage and other criteria. The only difference is the definition of the inter-cluster dissimilarity function where with HCC we use $dis^{CC}(.,.)$ defined in Eq. \ref{eq:distCC} (inspired from the cost function of flat correlation clustering). Other than this, the algorithmic procedure is consistent. 
Algorithm \ref{alg:hierarchical_CC} in Appendix \ref{Appendix:HCC_Alg} describes the pseudocode of the HCC algorithm.

\section{Feature Extraction from HCC}
As mentioned, HCC represents the relations between objects according to the way they join to form the hierarchy. In this section, we use this intuition and develop a data representation consistent with HCC.  
For this purpose, we adapt the methods in \cite{ShiehHA11,Shieh16916} and in particular \cite{ChehreghaniMLJ20} to our setting. Hereby, we 
first introduce distance functions over HCC, and then, investigate the embedding of such a distance function. This procedure leads to obtaining a set of features from HCC for each object which then can be used in the downstream task.

\subsection{Distance functions over HCC}

Given dendrogram $T$, each cluster $\mathbf v \in T$ represents the root of a dendrogram $T' \in \mathcal T_T$. $T'$ admits the properties of its root cluster, i.e., $level(T') = \max_{\mathbf u\in T'} level(\mathbf u) = level(\mathbf v)$ and $linkage(T') = \max_{\mathbf u\in T'} linkage(\mathbf u) = linkage(\mathbf v)$, since the root cluster has the maximum linkage and level among the clusters in $T'$. Hence, in this way, we define functions such as $level()$ and $linkage()$ for the dendrograms as well. 

The $linkage()$ function may seem to be a natural choice for defining a distance function over a HCC dendrogram.
Specifically, one can define the dendrogram-based distance function $\mathbf X_{ij}$ over dendrogram $T$ between $i, j\in \mathbf O$ as 
\begin{equation}
	\mathbf X_{ij} = \min linkage(T') \quad \text{s.t.} \quad i,j \in T',  \text{ and } T'\in \mathcal{T}_T.
\label{eq:linkage_hierarchy_dist}
\end{equation}

This choice corresponds to the linkage of the smallest cluster that includes both $i$ and $j$.
This in particular makes sense for the single linkage dendrogram, and it would be consistent with the tree-preserving embedding in \cite{ShiehHA11,Shieh16916}.
If the original dissimilarity matrix $\mathbf D$ contains negative values, then using  Proposition  \ref{thm:agg_shift}, one can sufficiently shift the pairwise dissimilarities to make all of them non-negative, without changing the structure of the dendrogram and the order of the clusters. Therefore, the conditions for a valid distance function including non-negativity still hold.

However, for the HCC dendrogram, the linkage function might not fulfill the conditions for a distance function. For example, consider a set of $n$ objects where all the pairwise similarities are $+1$, i.e., the dissimilarities are thus $-1$. Then, the linkage function will always return negative values which would violate the non-negativity condition of a valid distance function.
On the other hand, the HCC linkage $dis^{CC}(.,.)$ is \emph{not shift-invariant} (similar to the standard flat correlation clustering \cite{MortezaShift_MLj}) and we cannot use the shift trick in Proposition \ref{thm:agg_shift}. Let $\mathbf D^\alpha$ shows the shifted pairwise dissimilarities, i.e., $\mathbf D^\alpha_{i,j} = \mathbf D_{ij} + \alpha$. 
Then, $dis^{CC}(\mathbf u,\mathbf v)$ between two clusters $\mathbf u$ and $\mathbf v$ based on $\mathbf D^\alpha$ is given by
 \begin{eqnarray}\label{eq:hcc_linkage_shift}
    dis^{CC}(\mathbf u,\mathbf v) &=& \sum_{i \in \mathbf u, j \in \mathbf v} \mathbf D^\alpha_{i,j}
    = \sum_{i \in \mathbf u, j \in \mathbf v} (\mathbf D_{i,j} + \alpha)   \nonumber \\
    &=&
     \sum_{i \in \mathbf u, j \in \mathbf v} \mathbf D_{i,j}    + \alpha |\mathbf u||\mathbf v|\, .
  \end{eqnarray}
With $\alpha > 0$, this shift would induce a bias for the HCC linkage to choose imbalanced clusters. In other words, $dis^{CC}(.,.)$ is not shift-invariant and we cannot shift the pairwise dissimilarities in $\mathbf D$ to make them nonnegative.

Therefore, we consider another choice, i.e., the  $level()$ function used in \cite{ChehreghaniMLJ20}. It is nonnegative and satisfies the desired conditions. Then, $\mathbf X_{ij}$ is now computed by
\begin{equation}
	\mathbf X_{ij} = \min level(T') \quad \text{s.t. }  i,j\in T'  \text{ and } T'\in \mathcal{T}_T \, .
\label{eq:HCClevel}
\end{equation}

Intuitively, Eq. \ref{eq:HCClevel} selects the level of the smallest cluster/dendrogram that contains both $i$ and $j$. The lower the level at which the two objects join, the greater the similarity or proximity between them, indicating a closer relationship in the hierarchical clustering structure. In other words, a higher level in the dendrogram signifies a later fusion of the two objects, suggesting that they share fewer common characteristics compared to objects fused at lower levels.

\subsection{Embedding the HCC-based distances}

After applying the distance function in Eq.~\ref{eq:HCClevel}, we obtain an $n \times n$ matrix representing pairwise HCC-based distances among objects. It is usually preferred to obtain vector-based representations for objects rather than pairwise distances. Models like Gaussian Mixture Models (GMMs) which involve mixture density estimation (see, e.g., \cite{Titterington1985-uj}), can only be applied to vectors. 
 Additionally, working with vector-based data simplifies feature selection. Hence, it is desired to compute an embedding of the objects into a new space, so that their pairwise squared Euclidean distances in the new space match their pairwise distances obtained from the dendrogram.

The matrix of pairwise distances $\mathbf X$ obtained via Eq. \ref{eq:HCClevel} induces an \emph{ultrametric} \cite{Leclerc1981,ChehreghaniMLJ20}. 
The primary distinction between a \emph{metric} and an \emph{ultrametric} is that the \emph{addition} operation in the triangle inequality for a metric is replaced by a \emph{maximum} operation, i.e., with ultrametric we have

\begin{equation}
    \forall i,j,k: \mathbf X_{ij}  \le  \max(\mathbf X_{ik},\mathbf X_{kj}).
    \label{eq:ultrametric}
\end{equation}

The connection between ultrametric and trees is well-established in mathematics \cite{HUGHES2004148,MartnezPrez2008}. Here we instantiate it to our setting via making the argument in \cite{ChehreghaniMLJ20} more accurate.

\begin{figure}[th]
  \centering
  \includegraphics[width=0.4\textwidth]{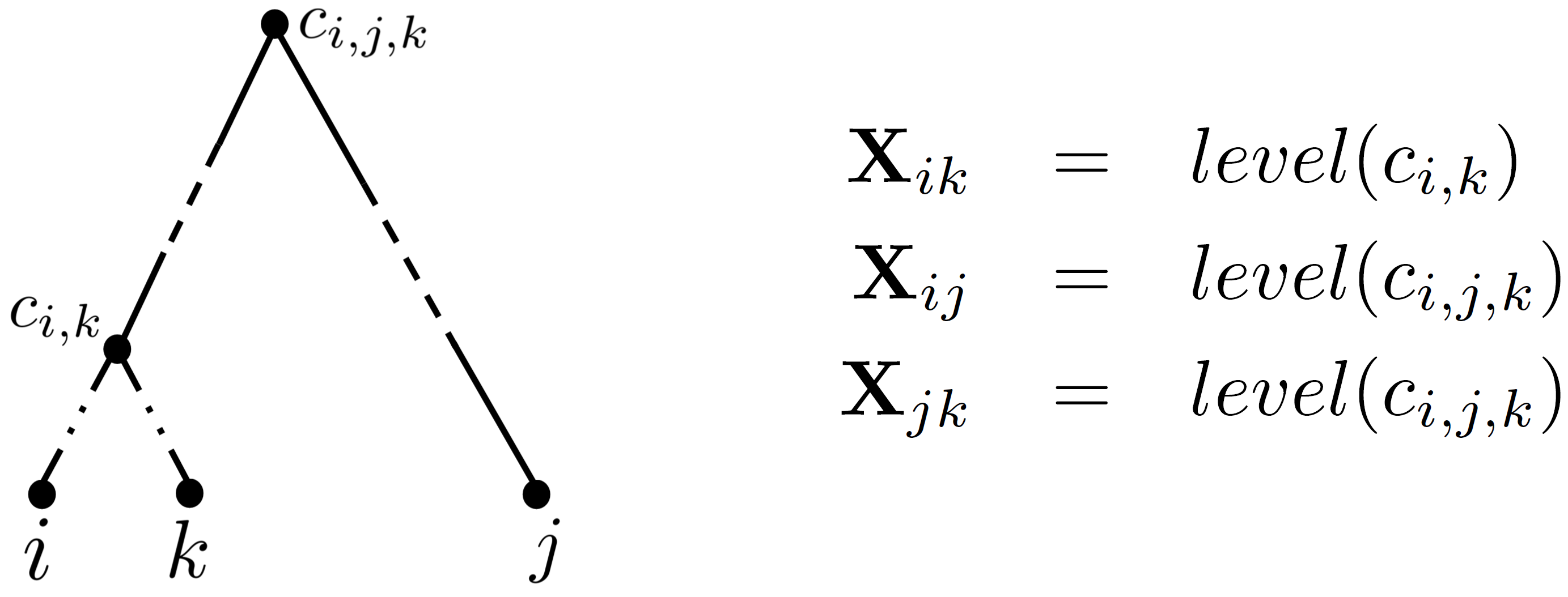}
  \caption{Illustration of ultrametric property of $\mathbf X$.}
  \label{Fig:ultrametric}
\end{figure}

It is evident that when $\mathbf X_{ij} \le \mathbf X_{ik}$, the inequality in Eq. \ref{eq:ultrametric} is satisfied. Conversely, if $\mathbf X_{ij} > \mathbf X_{ik}$, it implies that objects $i$ and $k$ are included in the same cluster (shown by $c_{i,k}$) before $i$ and $j$ join (to form cluster $c_{i,j}$). 
The bottom-up hierarchical clustering process then continues until $c_{i,j,k}$ is formed, encompassing all three objects $i,j,k$.
Notice that $i$ and $k$ join $j$ simultaneously via $c_{i,k}$.    In this case, according to Eqs. \ref{eq:HCClevel} and \ref{eq:ultrametric}, and the relationships illustrated in Figure \ref{Fig:ultrametric}, we conclude 
 \begin{equation}
   \mathbf X_{ij} = \mathbf X_{kj} \le \max(\mathbf X_{ik},\mathbf X_{kj}).
 \end{equation}

Ultrametric matrices, in turn, exhibit positive definiteness \cite{Fiedler1998, Varga1993OnSU}, and such positive definite matrices result in inducing an Euclidean embedding \cite{Schoenberg}. Thereby, after ensuring the existence of such an embedding, we can employ a proper method to compute it. Specifically, we use the method proposed in \cite{RePEc1938} known as \emph{multidimensional scaling} \cite{MDS}. This method proposes first centering $\mathbf X$ to obtain a Mercer kernel and then performing an eigenvalue decomposition. It works as follows.
\begin{enumerate}[leftmargin=4mm]
\item We center $\mathbf X$ by
    \begin{equation}
   	 \mathbf B\leftarrow -\frac{1}{2}
   (\mathbf{I}_n - \frac{1}{n}\mathbf{L}_n) \mathbf X (\mathbf{I}_n - \frac{1}{n}\mathbf{L}_n),
	\label{Eq:centering}
    \end{equation}
    where  $\mathbf{I}_n$ is an identity matrix of size $n\times n$ and $\mathbf{L}_n$ represents an $n\times n$ matrix filled entirely with ones. With this centering, the sum of both the rows and columns in matrix $\mathbf B$ becomes zero.
\item With applying the transformation in step 1, $\mathbf B$ becomes a positive semidefinite matrix, i.e., all the eigenvalues are nonnegative. Thus, we decompose $\mathbf B$ into its eigenbasis:
     \begin{equation}
       \mathbf B=\mathbf{Y}\mathbf{Z}\mathbf{Y}^{\texttt T},
     \end{equation}
     where $\mathbf{Y} = (\mathbf y_1,...,\mathbf y_n)$ represents the eigenvectors $\mathbf y_i$ and $\mathbf{Z}=\texttt{diag}(z_1,...,z_n)$ is a diagonal matrix of eigenvalues $z_1\geq...\geq z_l\geq z_{l+1}= 0 = ... = z_n$.
\item We calculate the $n\times l$ matrix $\mathbf A$ as the matrix of new data features:
    \begin{equation}
      \mathbf A=\mathbf{Y}_l(\mathbf{Z}_l)^{1/2},
    \end{equation}
    with $\mathbf{Y}_l=(\mathbf y_1,...,\mathbf y_l)$ and   $\mathbf{Z}_l=\texttt{diag}(z_1,..., z_l)$, where $l$ specifies the dimensionality of the new vectors.
\end{enumerate}

In the embedded space, the new dimensions are arranged based on their corresponding eigenvalues. One can opt to select only the most significant ones, rather than utilizing all of them. Therefore, computing such an embedding also offers the benefit of feature selection. 
 
We also note that many clustering methods can be written in matrix factorization form via for example spectral $K$-means \cite{DhillonGK04}. This induces an embedding and hence a set of relevant features. However, for general positive/negative similarity matrices no exact embedding might be feasible due to violating positive semidefiniteness. The method that we described here provides a solution to this challenge, i.e., enables us to extract features when the base pairwise similarities are positive and negative.  

\section{Correlation Clustering and Minimax Dissimilarities}

Finally, we study minimax dissimilarities for correlation clustering, a graph-based method that corresponds to constructing a hierarchical clustering. 

Given graph $\mathcal G(\mathbf O,\mathbf D)$, the minimax (MM) dissimilarity between  $i$ and $j$ is defined as
\begin{eqnarray}
	\mathbf D_{ij}^{MM} &=& \min_{p\in \mathcal P_{ij}(\mathcal G)}\max_{1\le l \le |p|-1}\mathbf D_{p(l)p(l+1)},
	\label{eq:PathStandard}
\end{eqnarray}
where $\mathcal P_{ij}(\mathcal G)$ is the set of all paths between $i$ and $j$ over $\mathcal G(\mathbf O, \mathbf D)$. Each path $p$ is specified by a sequence of object indices, i.e., $p(l)$
indicates the $l^{th}$ object on the path.

Minimax dissimilarities enable a clustering algorithm to capture the inherent patterns and manifolds in an unsupervised and nonparametric way by extracting the transitive connections \cite{FischerB03,Chehreghani20}.  For example, if object $i$ is similar to object $j$, $j$ is similar to $k$, and $k$ is similar to $l$, then the minimax dissimilarity between $i$ and $l$ will be small, even though their direct dissimilarity might be large. The reason is that minimax dissimilarity finds the connectivity path $i \to j \to \cdots \to k \to l$ and connects $i$ and $l$ via this path. This property is helpful in finding elongated clusters and manifolds of arbitrary shapes in an unsupervised way.

Minimax dissimilarities have been so far solely used with nonnegative pairwise dissimilarities. 
In the case of possible negative dissimilarities, we may use a trick similar to Proposition \ref{thm:agg_shift}. As shown in Lemma \ref{lemma:minimax_invariant}, minimax paths are invariant w.r.t. the shift of the pairwise dissimilarities.

\begin{lemma}
Consider graphs $\mathcal G(\mathbf O,\mathbf D)$  and $\mathcal G^\alpha(\mathbf O,\mathbf D^\alpha)$, where the pairwise dissimilarities (edge weights) in $\mathcal G^\alpha(\mathbf O,\mathbf D^\alpha)$ are shifted by constant  $\alpha$, i.e., $\mathbf D^\alpha_{i,j} = \mathbf D_{i,j} + \alpha$. Then, the minimax paths between every pair of objects $i$ and $j$ are identical on graphs $\mathcal G(\mathbf O,\mathbf D)$ and $\mathcal G^\alpha(\mathbf O,\mathbf D^\alpha)$.
\label{lemma:minimax_invariant}
\end{lemma}

All the proofs are in Appendix \ref{Appendix:Proofs}.
Hence, given a dissimilarity matrix $\mathbf D$, one can subtract $\alpha := min(\mathbf D)$ from all the elements to obtain $\mathbf D^\alpha$. Then, the minimax dissimilarities can be computed from $\mathcal G^\alpha(\mathbf O,\mathbf D^\alpha)$.
After computing the minimax dissimilarities from $\mathcal G^\alpha$, we may add $\alpha$ to all the pairwise minimax dissimilarities. We can obtain the  minimax similarities $\mathbf S^{MM}_{ij}$ via  $\mathbf S^{MM}_{ij} = - \mathbf D^{MM}_{ij}$, if needed.

We demonstrate that for correlation clustering there exists a simpler method to calculate the minimax dissimilarities intended for use in correlation clustering. 
According to Theorem \ref{thm:CC_MM}, performing correlation clustering on minimax dissimilarities can be achieved in polynomial time via computing the  connected components of the unweighted graph
 $\mathcal G(\mathbf O,\mathbf S')$, where the similarity matrix  $\mathbf S'$ is obtained by
 $\mathbf S'_{ij}=  1$  if  $\mathbf S_{ij} >0$, and  $\mathbf S'_{ij}=  0$ otherwise.

\begin{theorem}
The optimal clusters of the correlation clustering on graph $\mathcal G(\mathbf O, \mathbf S^{MM})$ are equal to the connected components of graph $\mathcal G(\mathbf O, \mathbf S')$.
\label{thm:CC_MM}
\end{theorem}

As mentioned, correlation clustering on an arbitrary similarity matrix $\mathbf S$ is NP-hard \cite{BansalBC04,DemaineEFI06}.
Therefore, using minimax (dis)similarities with correlation clustering not only helps for extracting elongated complex patterns, but also yields a significant reduction in the computational complexity, i.e., from NP-hardness to a polynomial runtime.

Among the approximate algorithms proposed for correlation clustering on complete graphs with discrete weights, the method in \cite{AilonCN08} provides a  3-factor approximation. with a randomly selected unclustered object in the graph, this method greedily finds the object's positive neighbors (those with similarity $+1$) to form a new cluster. Then, it repeats this procedure for the remaining objects. One can conclude that in the optimal solution of correlation clustering on graph $\mathcal G(\mathbf O, \mathbf S^{MM})$, only the positive neighbors of an object will be in the same cluster as the object is, i.e., interestingly the 3-factor approximation algorithm in \cite{AilonCN08} becomes exact when applied to $\mathcal G(\mathbf O, \mathbf S^{MM})$ (Theorem \ref{thm:CC_approx_MM}).

\begin{theorem}
Assume the edge weights of graph $\mathcal G(\mathbf O, \mathbf S)$  are either $+1$ or $-1$. Then, the approximate algorithm in \cite{AilonCN08} is exact when applied to the minimax similarities, i.e., to graph $\mathcal G(\mathbf O, \mathbf S^{MM})$.
\label{thm:CC_approx_MM}
\end{theorem}

\section{Experiments}
In this section, we describe our experimental results on various datasets. We compare our methods with single linkage (SL), complete linkage (CL) and average linkage (AL).\footnote{
Some criteria, e.g.  \emph{centroid},  \emph{median} and \emph{Ward}  compute a representative for each cluster and then compute the inter-cluster dissimilarities by the distances between the representatives. Computing such representatives might not be feasible for possibly negative pairwise dissimilarities.
 Thus, we do not consider them.} As mentioned, there are several improvements over these basic methods. However, such contributions are orthogonal to our contribution. Moreover, it is unclear how such improvements can be extended to the dissimilarities that can be both positive and negative. Thus, we limit our baselines to these methods which as mentioned in Proposition \ref{thm:agg_shift}, are applicable to such data. 

In our studies, we have access to the true labels. Therefore, consistent with several previous studies on hierarchical clustering, e.g. \cite{BateniBDHKLM17,Ah-Pine18,DhulipalaELMS21},  we evaluate the results according to the following criteria: 
i) Normalized Mutual Information (MI) \cite{Vinh:2010} that measures the mutual information between the true and the estimated clustering solutions, and
ii) Normalized Rand score (Rand) \cite{hubert1985comparing} that  obtains  the similarity between the two solutions.
We do not use the labels to infer the clustering solution, they are only used for evaluation. Therefore, we are still in the unsupervised setting where the ground-truth labels play the role of an external evaluator.

In Appendix \ref{Appendix:AddExperiments}, we describe additional experimental results, in particular on datasets from other domains.

\begin{figure*}[ht!]
    \centering
    \hspace{-19mm}
    \begin{subfigure}{0.20\linewidth}
        \includegraphics[width=1\textwidth]{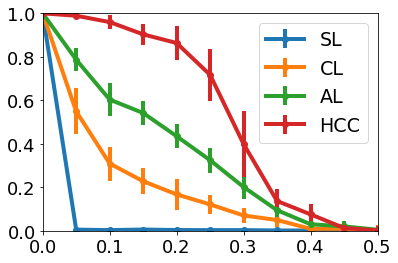}
        \caption{[\emph{Breast Tissue}]}
        \label{fig:Breast_Tissue_MI}
    \end{subfigure}
    \hspace{-5mm}
    \begin{subfigure}{0.20\linewidth}
        \includegraphics[width=1\textwidth]{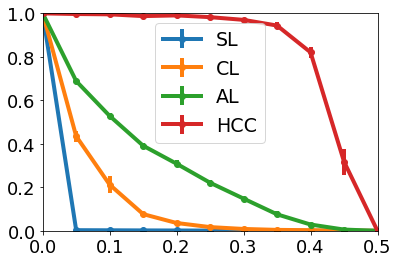}
        \caption{\emph{Cardio}}
        \label{fig:Cardiotocography_MI}
    \end{subfigure}
    \hspace{-5mm}
    \begin{subfigure}{0.20\linewidth}
        \includegraphics[width=1\textwidth]{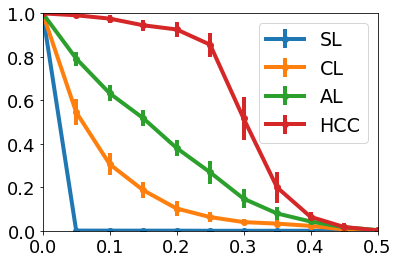}
        \caption{\emph{Image Segmentation}}
    \label{fig:Image_Segmentation_MI}
    \end{subfigure}
    \hspace{-5mm}
    \begin{subfigure}{0.20\linewidth}
        \includegraphics[width=1\textwidth]{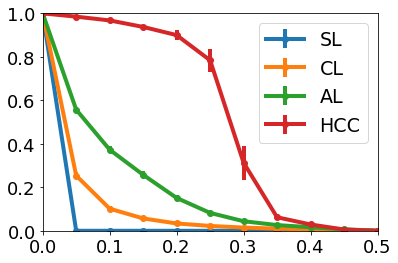}
        \caption{\emph{ISOLET}}
        \label{fig:ISOLET5_MI}
    \end{subfigure}
    \hspace{-5mm}
    \begin{subfigure}{0.20\linewidth}
        \includegraphics[width=1\textwidth]{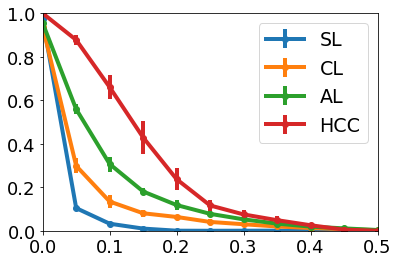}
        \caption{\emph{Leaf}}
        \label{fig:Leaf_MI}
    \end{subfigure}
    \hspace{-5mm}
    \begin{subfigure}{0.20\linewidth}
        \includegraphics[width=1\textwidth]{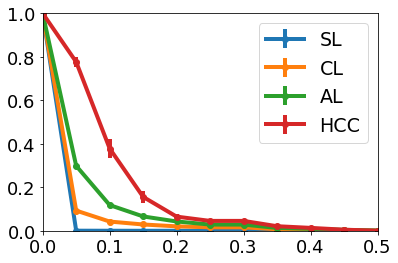}
        \caption{\emph{One-Hundred Plant}}
        \label{fig:One_hundred_plant_MI}
    \end{subfigure}
    \hspace{-25mm}
    \caption{MI score of different hierarchical clustering methods applied to  UCI datasets, where x-axis shows the parameter $\eta$.}
    \label{fig:UCI_hierarchical_MI}
\end{figure*}

\begin{table*}[thb!]
\begin{adjustwidth}{0cm}{0cm}
\caption{Performance of different tree preserving embedding methods on UCI datasets applied with GMM.
}
\centering
\begin{tabular}{|| c || c c || c c || c c || c c || c c || c c ||} 
\hline\hline 
 &\multicolumn{2}{c||}{\emph{Breast Tissue}}&\multicolumn{2}{c||}{\emph{Cardiotocography}}&\multicolumn{2}{c||}{\emph{Image Segm.}}&\multicolumn{2}{c||}{\emph{ISOLET}}&\multicolumn{2}{c||}{\emph{Leaf}}&\multicolumn{2}{c||}{\emph{One-Hun. Plant}}\\
 \hline
 method & MI & Rand & MI & Rand & MI & Rand & MI & Rand  & MI & Rand & MI & Rand \\ 
\hline 
SL                & 0.006 & 0.008 & 0.007 & 0.007 & 0.008 & 0.001 & 0.009 & 0.016 & 0.008 & 0.003 & 0.015 & 0.020 \\
SL+GMM     & 0.093 & 0.077 & 0.120 & 0.135 & 0.239 & 0.250 & 0.192 & 0.174 & 0.155 & 0.161 & 0.083 & 0.077\\
\hline
CL               & 0.227  & 0.166 & 0.077 & 0.056 & 0.187 & 0.125 & 0.057 & 0.017 & 0.081 & 0.038 & 0.029 & 0.008 \\
CL+GMM    & 0.251 & 0.171 &  0.081 & 0.060 & 0.201 & 0.154 & 0.061 & 0.043 & 0.140 & 0.129 & 0.054 & 0.049 \\
\hline
AL               & 0.542  & 0.519 & 0.391 & 0.479 & 0.518 & 0.495 & 0.257 & 0.165 & 0.181 & 0.106 & 0.066 & 0.023 \\
AL+GMM    & 0.550 & 0.513 & 0.422 & 0. 463 & 0.522 & 0.501 & 0.240 & 0.179 & 0.152 & 0.143 & 0.081 & 0.065 \\
\hline
HCC            & 0.903 & 0.900 & \bf{0.987} & \bf{0.994} & 0.945 & 0.943 & 0.938 & \bf{0.918} & 0.429 & 0.373 & 0.159 & 0.104 \\
HCC+GMM & \bf{0.914} & \bf{0.911} & 0.979 & 0.974 & \bf{0.960} & \bf{0.966} & \bf{0.941} & 0.917 & \bf{0.462} & \bf{0.401} & \bf{0.183} & \bf{0.217} \\
\hline
\end{tabular}
\label{table:UCI_embedding_GMM} 
\end{adjustwidth}
\end{table*}

\subsection{HCC on UCI data}
We first investigate the hierarchical correlation clustering on the following six UCI datasets  
\cite{UCI_Datasets}.
(i) \emph{Breast Tissue}: includes electrical impedance measurements of freshly excised $106$  tissue samples from the breast. The number of clusters is $6$. 
(ii) \emph{Cardiotocography}: contains $2126$ measurements of fetal heart rate and uterine contraction  features on cardiotocograms in $10$ clusters. 
(iii) \emph{Image Segmentation}: contains $2310$ samples from images of $7$ outdoor clusters. 
(iv) \emph{ISOLET}:  $7797$ samples consisting of spoken attributes of different letters ($26$ clusters). 
(v) \emph{Leaf}: $340$ images of leaf specimens originating from $40$ different plant species (clusters) each described by $16$ attributes. 
(vi) \emph{One-Hundred Plant}: $1600$ samples of leafs each described by $64$ features, from in total $100$ types (clusters). 
%
The ground-truth labels are shown by $\mathbf c^*$, i.e., $\mathbf c^*_i$ shows the true label for object $i$. We assume an oracle reveals the pairwise similarities $\mathbf S$ according to the (flip) noise parameter $\eta$. 
If $\mathbf c^*_i=\mathbf c^*_j$ then $\mathbf S_{i,j} = \mathcal{U}(0,1)$ with probability $1-\eta$ and $\mathbf S_{i,j} = \mathcal{U}(-1,0)$ with probability $\eta$.  If $\mathbf c^*_i\ne \mathbf c^*_j$ then $\mathbf S_{i,j} = \mathcal{U}(-1,0)$ with probability $1-\eta$ and $\mathbf S_{i,j} = \mathcal{U}(0,1)$ with probability $\eta$. The function $ \mathcal{U}(.,.)$ returns a uniform random number within the specified range. For each $\eta$ we repeat the experiments $20$ times and report the average results. This setup provides a systematic approach to study the robustness of various methods to noise. 

Figure \ref{fig:UCI_hierarchical_MI}   shows the results for different datasets as a function of the noise level $\eta$ w.r.t.  MI.
Rand scores shown in Figure \ref{fig:UCI_hierarchical_Rand} in Appendix \ref{Appendix:AddExperiments} exhibit a consistent behavior.
We observe that among different methods, HCC performs significantly better and produces more robust clusters w.r.t. the noise parameter $\eta$. The results on \emph{Leaf} and \emph{One-Hundred Plant} are worse with all the methods. The reason is that these datasets are complex, having many clusters (respectively, $40$ and $100$ clusters) and fairly a small number of objects per cluster.

\subsection{Tree preserving embedding on UCI data}
In the following, we investigate tree preserving embedding and feature extraction. 
After computing the embeddings from different hierarchical clustering methods, we apply Gaussian Mixture Model (GMM) to the extracted features and evaluate the final clustering using the ground-truth solution.
This kind of embedding enables us to apply methods like  GMM to positive and negative pairwise similarities, a task that was not possible before. Since the extracted features appear in the form of vectors, thus, the final clustering method is not limited to GMM, and in principle, any numerical clustering can benefit from this embedding.

Table \ref{table:UCI_embedding_GMM} 
shows the results for different UCI datasets. We observe that the embeddings obtained by HCC (i.e., `HCC+GMM') yield significantly better results compared to the embeddings from other methods (i.e., `SL+GMM', `CL+GMM' and `AL+GMM').  It is worth noting that the results of embeddings (e.g., `HCC+GMM') typically surpass the results obtained from the hierarchical clustering alone (e.g., `HCC'). This observation supports the idea that employing HCC to compute an embedding (for extracting new features) for a clustering method such as GMM may be advantageous, yielding superior results compared to using HCC exclusively for clustering purposes. This verifies why tree preserving embedding can be effective in general.

\subsection{Experiments on Fashion-MNIST}
Next, we investigate HCC and tree-preserving embedding on two randomly selected subsets of Fashion-MNIST dataset \cite{abs-1708-07747}. Fashion MNIST consists of $28\times 28$ images of Zalando's articles. Each subset consists of $5,000$ samples/objects, where we compute the pairwise cosine similarities between them and then apply different methods. Table \ref{table:image_data} shows the performance of different methods on these datasets. We observe that, consistent with the previous experiments, both `HCC' and `HCC+GMM' yield improving the results compared to the baselines. Furthermore, employing HCC to compute intermediate features for GMM (i.e., `HCC+GMM') achieves higher scores compared to using `HCC' alone for generating final clusters.

\begin{table}[t]
\caption{Performance of different methods on MNIST and Fashion MNIST datasets. The embeddings by HCC yield better results.
}
\centering 
\begin{tabular}{|| c || c c | c c ||} 
\hline\hline 
 \multicolumn{1}{||c||}{} & \multicolumn{2}{c|}{Fashion MNIST 1} & \multicolumn{2}{c||}{Fashion MNIST 2} \\
 \hline 
 method & MI & Rand & MI & Rand \\
\hline 
SL     & 0.322 & 0.206 & 0.241 & 0.196 \\
SL+GMM & 0.411 & 0.335 & 0.384 & 0.340 \\
\hline
CL     & 0.403 & 0.293 & 0.546 & 0.379 \\
CL+GMM & 0.478 & 0.426 & 0.574 & 0.362 \\
\hline
AL     & 0.464 & 0.468 & 0.602 & 0.534 \\
AL+GMM & \bf{0.608} & 0.551 & 0.647 & 0.553 \\
\hline
HCC    & 0.499 & 0.475 & 0.666 & \bf{0.557} \\
HCC+GMM & 0.581 & \bf{0.586} & \bf{0.693} & 0.548 \\
\hline 
\end{tabular}
\label{table:image_data} 
\end{table}

\begin{figure}[thb]
    \centering
    \begin{subfigure}{0.15\textwidth}
        \includegraphics[width=1\textwidth]{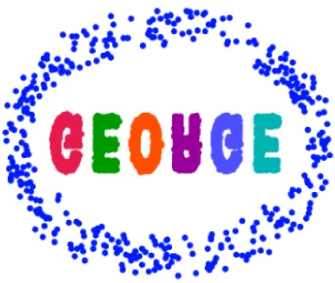}
        \caption{DS1 \cite{CLUTO,Peng2022}}
        \label{fig:NatCommMnimax}
    \end{subfigure}
    \hspace{7mm}
    \begin{subfigure}{0.18\textwidth}
        \includegraphics[width=1\textwidth]{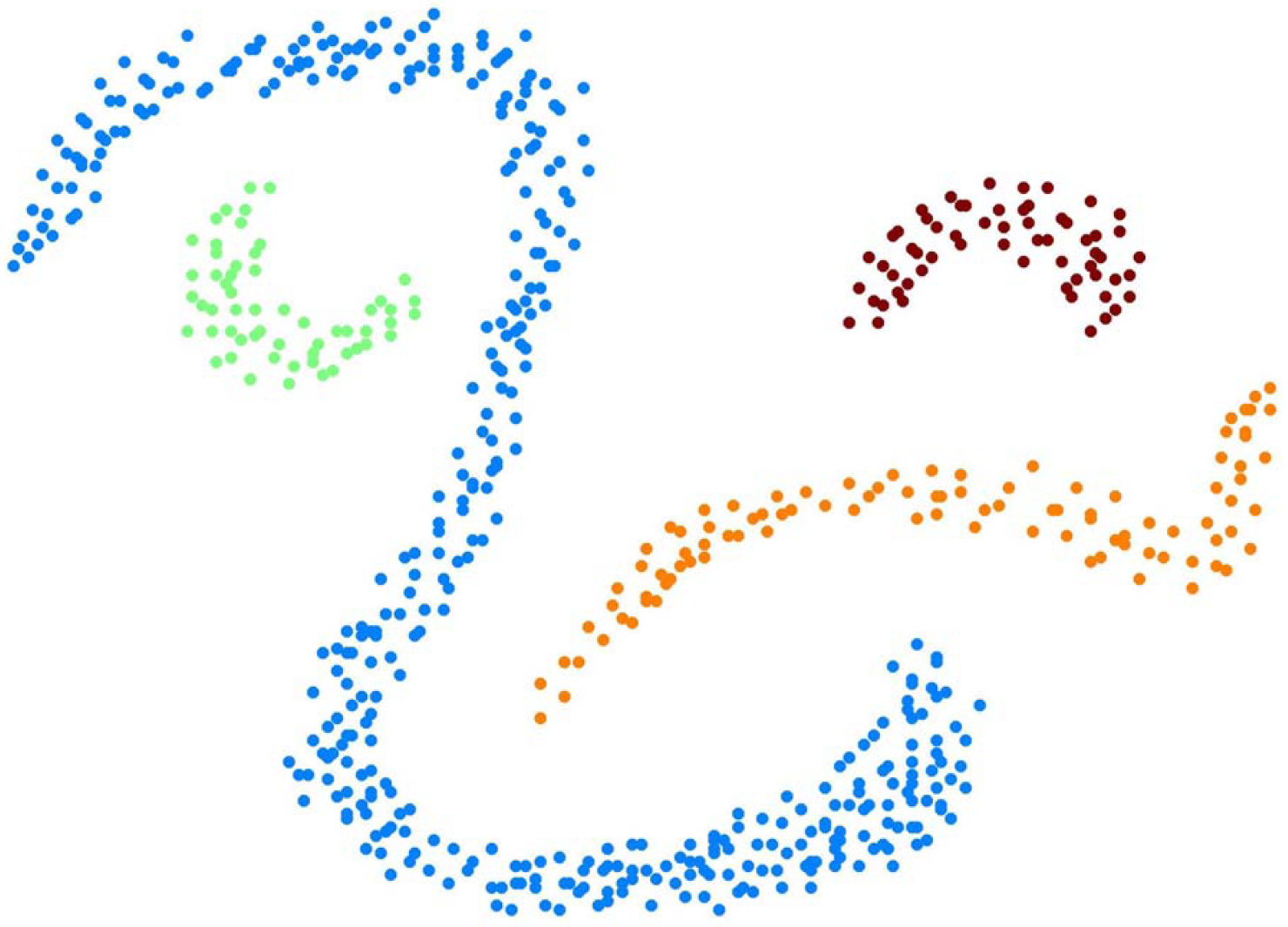}
        \caption{DS2 \cite{ChengZHWY19}}
        \label{fig:SMinimax}
    \end{subfigure}
    \\
    \begin{subfigure}{0.13\textwidth}
        \includegraphics[width=1\textwidth]{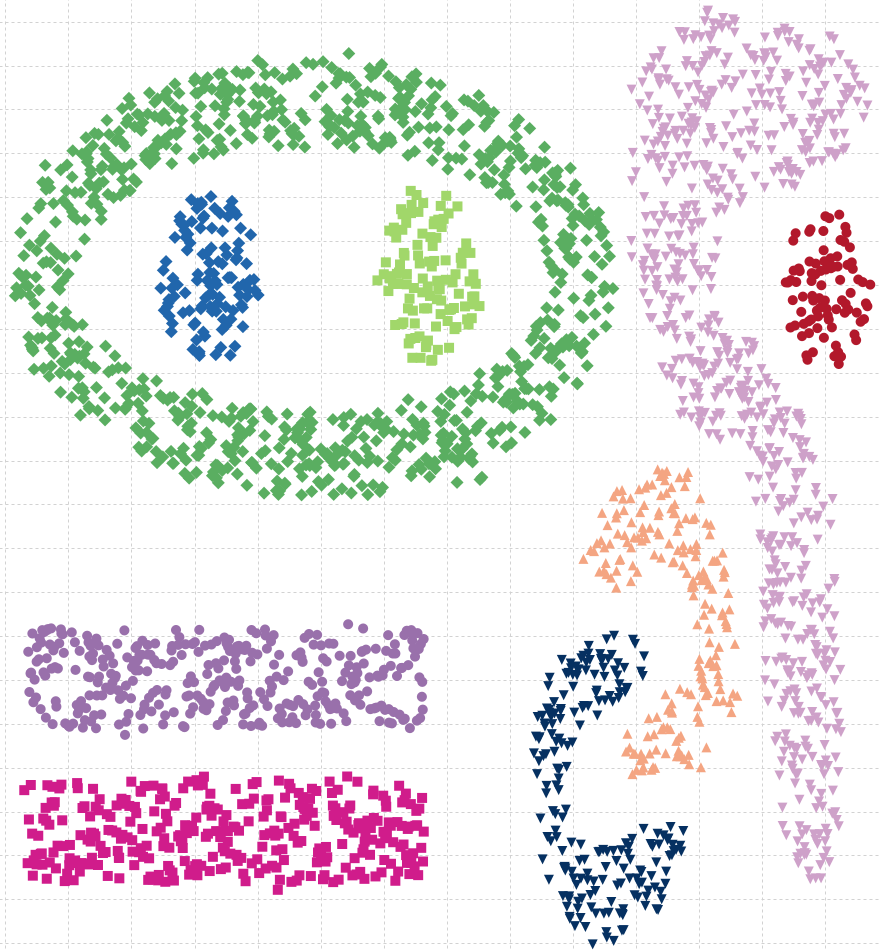}
        \caption{DS3 \cite{Clustering_Datasets}}
        \label{fig:complex9Minimax}
    \end{subfigure}
    \hspace{10mm}
    \begin{subfigure}{0.12\textwidth}
        \includegraphics[width=1\textwidth]{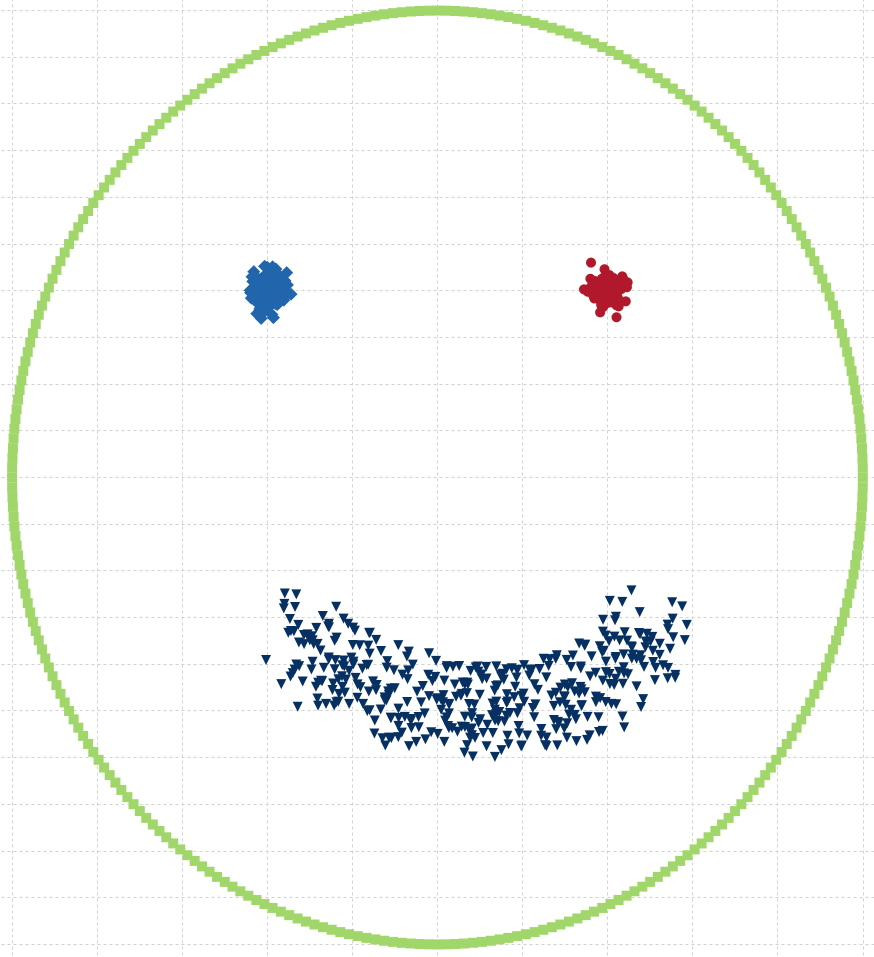}
        \caption{DS4 \cite{Clustering_Datasets}}
        \label{fig:smile2Minimax}
    \end{subfigure}
    \caption{The datasets with arbitrarily shaped clusters, where \emph{`minimax + correlation clustering'} acheives perfect clustering.
    }
    \label{fig:Minimax_Visual}
\end{figure}

\begin{figure}[th]
  \centering
  \includegraphics[width=0.3\textwidth]{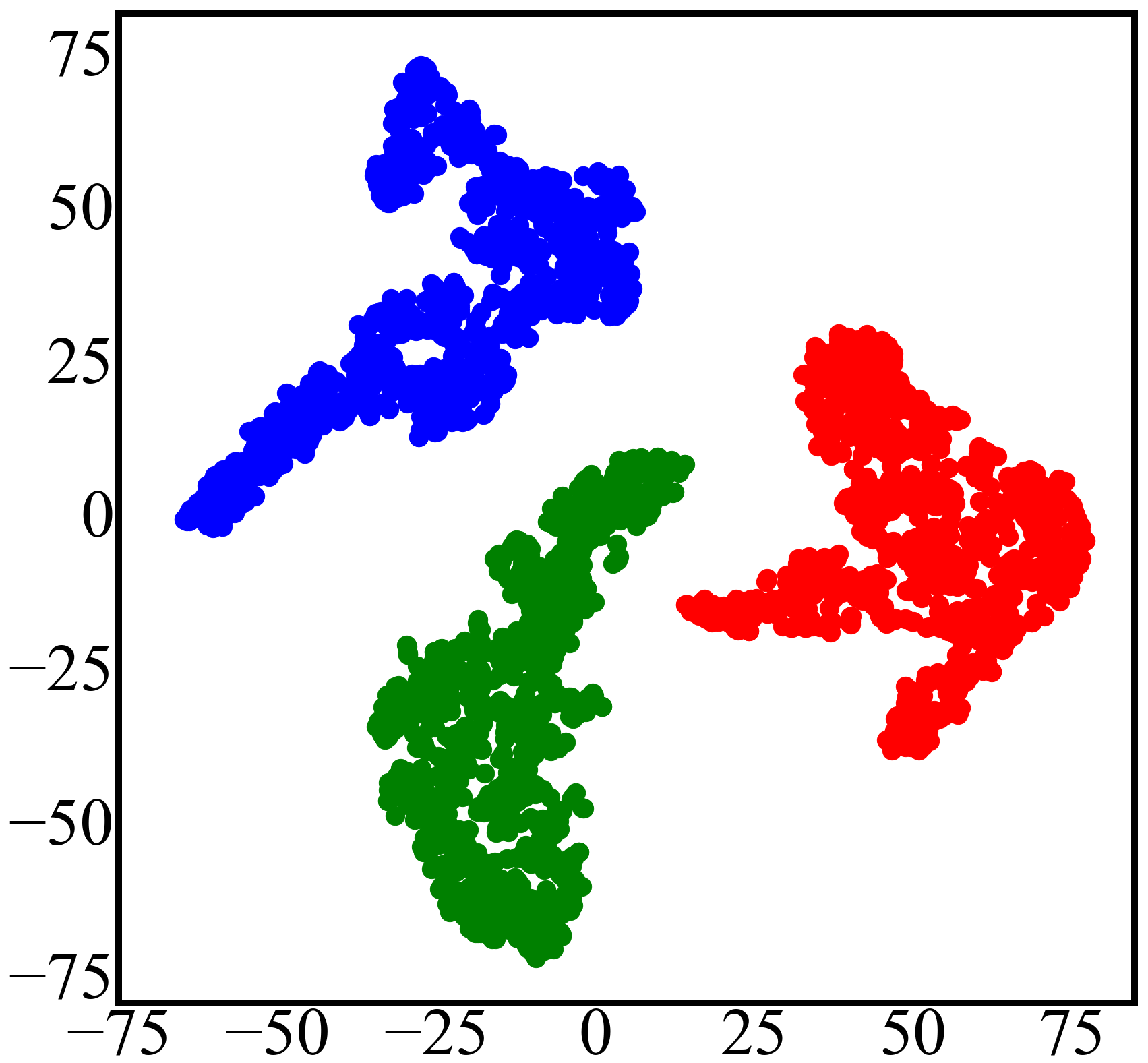}
  \caption{Embeddings of vehicle motion trajectories computed via dynamic time wrapping and t-SNE \cite{HoseiniRC21,DemetriouARC23}.}
  \label{Fig:set5_1024_data}
\end{figure}

\subsection{Correlation clustering with minimax dissimilarities}

Finally, we investigate the use of minimax dissimilarities with correlation clustering. As mentioned, minimax dissimilarities are especially useful for extracting elongated and complex patterns in an unsupervised way. Thereby, we apply \emph{`minimax + correlation clustering'} to a number of datasets with visually elongated and arbitrarily shaped clusters, and compare the results with \emph{`correlation clustering'} alone. The datasets are shown in Figure \ref{fig:Minimax_Visual}. For each dataset,  we simply construct the K-nearest neighbor graph (using the ordinary Euclidean distance with a typical K such as 3). The edges in the nearest neighborhood receive a positive weight (e.g., +1), and a negative weight (e.g., -1) otherwise. We then apply either \emph{`minimax + correlation clustering'} or \emph{`correlation clustering'} to the resultant graph. We observe that for all the datasets, \emph{`minimax + correlation clustering'} yields the perfect clustering, i.e., MI and Rand are equal to 1. Whereas with \emph{`correlation clustering'} alone these scores are very low.  
We acknowledge the existence of other clustering methods, such as DBSCAN \cite{EsterKSX96},  capable of achieving perfect clustering on these datasets. 
However, these methods often involve \emph{crucial hyperparameters}, the tuning of which can be challenging in unsupervised learning. 
To our knowledge, in addition to computational and theoretical benefits, \emph{`minimax + correlation clustering'} stands out as the sole method capable of achieving perfect results on datasets with elongated clusters of arbitrary shapes, and fulfills the following two promises: i) \emph{it eliminates the need to fix critical hyperparameters, a task often intricate in unsupervised learning}, and ii) \emph{it automatically determines the correct number of clusters without requiring prior fixing}. 

In the following, we consider the interesting application of clustering vehicle motion trajectories for ensuring safety in self-driving \cite{HoseiniRC21}. In this application, $1,024$ trajectories consisting of drive-by left, drive-by right and cut-in types are prepared, where some of them are collected in real-world and some others are generated using Recurrent Auto-Encoder GAN \cite{DemetriouARC23}. Next, dynamic time wrapping \cite{Bellman1959adaptive} and t-SNE \cite{vandermaaten08a} are employed to map the temporal data onto a two-dimensional space, as illustrated in Figure \ref{Fig:set5_1024_data} \cite{HoseiniRC21,DemetriouARC23}. We then apply \emph{`minimax + correlation clustering'} to this data and compare it with \emph{`correlation clustering'}. Similar to the datasets in Figure \ref{fig:Minimax_Visual}, \emph{`minimax + correlation clustering'} achieves perfect clustering with MI = Rand = 1 and accurately determines the correct number of clusters.

\section{Conclusion}
We proposed a new hierarchical clustering method, called HCC, suitable when the (dis)similarities can be positive or negative. This method is consistent with the general algorithmic procedure for agglomerative clustering and only differs in the way the inter-cluster dissimilarity function is defined.  We then considered embedding the HCC dendrograms, which provides extracting useful features to apply for example GMM for clustering positive and negative similarities. In the following,
we studied the use of minimax dissimilarities
with correlation clustering and showed that it yields reduction in the computational complexity, in addition to a possibility for extracting elongated manifolds. Finally, we demonstrated the effectiveness of the methods on several datasets  in different settings.

\section*{Acknowledgement}
The work of Morteza Haghir Chehreghani was partially supported by the Swedish
Research Council VR (grant number 2023-04809) and the Wallenberg AI, Autonomous Systems and Software Program (WASP) funded by the Knut and Alice Wallenberg Foundation.

\newpage

{\small
\bibliographystyle{ieee_fullname}
\bibliography{main}

\begin{thebibliography}{10}\itemsep=-1pt

\bibitem{Clustering_Datasets}
Clustering datasets.
\newblock \url{https://github.com/milaan9/Clustering-Datasets/}.

\bibitem{AchtertBKZ06}
Elke Achtert, Christian B{\"{o}}hm, Peer Kr{\"{o}}ger, and Arthur Zimek.
\newblock Mining hierarchies of correlation clusters.
\newblock In {\em 18th International Conference on Scientific and Statistical
  Database Management, {SSDBM}}, pages 119--128, 2006.

\bibitem{AgrawalHKMT09}
Rakesh Agrawal, Alan Halverson, Krishnaram Kenthapadi, Nina Mishra, and
  Panayiotis Tsaparas.
\newblock Generating labels from clicks.
\newblock In {\em International Conference on Web Search and Web Data Mining,
  {WSDM}}, pages 172--181, 2009.

\bibitem{Ah-Pine18}
Julien Ah{-}Pine.
\newblock An efficient and effective generic agglomerative hierarchical
  clustering approach.
\newblock {\em J. Mach. Learn. Res.}, 19:42:1--42:43, 2018.

\bibitem{AilonCN08}
Nir Ailon, Moses Charikar, and Alantha Newman.
\newblock Aggregating inconsistent information: Ranking and clustering.
\newblock {\em J. {ACM}}, 55(5):23:1--23:27, 2008.

\bibitem{Balcan:2014:RHC}
Maria-Florina Balcan, Yingyu Liang, and Pramod Gupta.
\newblock Robust hierarchical clustering.
\newblock {\em J. Mach. Learn. Res.}, 15(1):3831--3871, 2014.

\bibitem{BansalBC02}
Nikhil Bansal, Avrim Blum, and Shuchi Chawla.
\newblock Correlation clustering.
\newblock In {\em 43rd Symposium on Foundations of Computer Science {FOCS}},
  2002.

\bibitem{BansalBC04}
Nikhil Bansal, Avrim Blum, and Shuchi Chawla.
\newblock Correlation clustering.
\newblock {\em Machine Learning}, 56(1-3):89--113, 2004.

\bibitem{BateniBDHKLM17}
MohammadHossein Bateni, Soheil Behnezhad, Mahsa Derakhshan, MohammadTaghi
  Hajiaghayi, Raimondas Kiveris, Silvio Lattanzi, and Vahab~S. Mirrokni.
\newblock Affinity clustering: Hierarchical clustering at scale.
\newblock In {\em Advances in Neural Information Processing Systems (NIPS)},
  pages 6864--6874, 2017.

\bibitem{Bellman1959adaptive}
R. Bellman and R. Kalaba.
\newblock On adaptive control processes.
\newblock {\em Automatic Control, IRE Transactions on}, 4(2):1--9, 1959.

\bibitem{Bruynooghe1977}
Michel Bruynooghe.
\newblock Méthodes nouvelles en classification automatique de données
  taxinomiques nombreuses.
\newblock {\em Statistique et analyse des données}, 2(3):24--42, 1977.

\bibitem{ChakrabartiKP08}
Deepayan Chakrabarti, Ravi Kumar, and Kunal Punera.
\newblock A graph-theoretic approach to webpage segmentation.
\newblock In {\em International Conference on World Wide Web, {WWW}}, pages
  377--386, 2008.

\bibitem{CGW03}
Moses Charikar, Venkatesan Guruswami, and Anthony Wirth.
\newblock Clustering with qualitative information.
\newblock In {\em 44th Symposium on Foundations of Computer Science (FOCS)},
  pages 524--533, 2003.

\bibitem{Chehreghani17AAAI}
Morteza~Haghir Chehreghani.
\newblock Classification with minimax distance measures.
\newblock In {\em Proceedings of the Thirty-First {AAAI} Conference on
  Artificial Intelligence}, pages 1784--1790, 2017.

\bibitem{ChehreghaniICDM17}
Morteza~Haghir Chehreghani.
\newblock Clustering by shift.
\newblock In {\em {IEEE} International Conference on Data Mining (ICDM)}, pages
  793--798, 2017.

\bibitem{Chehreghani20}
Morteza~Haghir Chehreghani.
\newblock Unsupervised representation learning with minimax distance measures.
\newblock {\em Mach. Learn.}, 109(11):2063--2097, 2020.

\bibitem{ChehreghaniIJCNN21}
Morteza~Haghir Chehreghani.
\newblock Reliable agglomerative clustering.
\newblock In {\em International Joint Conference on Neural Networks, {IJCNN}},
  pages 1--8. {IEEE}, 2021.

\bibitem{MortezaShift_MLj}
Morteza~Haghir Chehreghani.
\newblock Shift of pairwise similarities for data clustering.
\newblock {\em Mach. Learn.}, 112(6):2025--2051, 2023.

\bibitem{ChehreghaniAC08}
Morteza~Haghir Chehreghani, Hassan Abolhassani, and Mostafa~Haghir Chehreghani.
\newblock Improving density-based methods for hierarchical clustering of web
  pages.
\newblock {\em Data Knowl. Eng.}, 67(1):30--50, 2008.

\bibitem{ChehreghaniAISTATS12}
Morteza~Haghir Chehreghani, Alberto~Giovanni Busetto, and Joachim~M. Buhmann.
\newblock Information theoretic model validation for spectral clustering.
\newblock In {\em International Conference on Artificial Intelligence and
  Statistics (AISTATS)}, 2012.

\bibitem{ChehreghaniMLJ20}
Morteza~Haghir Chehreghani and Mostafa~Haghir Chehreghani.
\newblock Learning representations from dendrograms.
\newblock {\em Mach. Learn.}, 109(9-10):1779--1802, 2020.

\bibitem{ChengZHWY19}
Dongdong Cheng, Qingsheng Zhu, Jinlong Huang, Quanwang Wu, and Lijun Yang.
\newblock A novel cluster validity index based on local cores.
\newblock {\em {IEEE} Trans. Neural Networks Learn. Syst.}, 30(4):985--999,
  2019.

\bibitem{Cochez:2015:TTA}
Michael Cochez and Hao Mou.
\newblock Twister tries: Approximate hierarchical agglomerative clustering for
  average distance in linear time.
\newblock In {\em International Conference on Management of Data (ACM SIGMOD)},
  pages 505--517, 2015.

\bibitem{Cohen-Addad:2018:HCO}
Vincent Cohen-Addad, Varun Kanada, Frederik Mallmann-Trenn, and Claire Mathieu.
\newblock Hierarchical clustering: Objective functions and algorithms.
\newblock In {\em SODA}, pages 378--397, 2018.

\bibitem{NIPS2017_7200}
Vincent Cohen-Addad, Varun Kanade, and Frederik Mallmann-Trenn.
\newblock Hierarchical clustering beyond the worst-case.
\newblock In {\em Advances in Neural Information Processing Systems (NIPS)},
  pages 6201--6209. 2017.

\bibitem{DemaineEFI06}
Erik~D. Demaine, Dotan Emanuel, Amos Fiat, and Nicole Immorlica.
\newblock Correlation clustering in general weighted graphs.
\newblock {\em Theor. Comput. Sci.}, 361(2-3):172--187, 2006.

\bibitem{DemetriouARC23}
Andreas Demetriou, Henrik Alfsv{\aa}g, Sadegh Rahrovani, and Morteza~Haghir
  Chehreghani.
\newblock A deep learning framework for generation and analysis of driving
  scenario trajectories.
\newblock {\em {SN} Comput. Sci.}, 4(3):251, 2023.

\bibitem{DhillonGK04}
Inderjit~S. Dhillon, Yuqiang Guan, and Brian Kulis.
\newblock Kernel k-means: spectral clustering and normalized cuts.
\newblock In {\em Tenth {ACM} International Conference on Knowledge Discovery
  and Data Mining, {SIGKDD}}, pages 551--556, 2004.

\bibitem{DhulipalaELMS21}
Laxman Dhulipala, David Eisenstat, Jakub Lacki, Vahab~S. Mirrokni, and Jessica
  Shi.
\newblock Hierarchical agglomerative graph clustering in nearly-linear time.
\newblock In {\em International Conference on Machine Learning (ICML)}, 2021.

\bibitem{EsterKSX96}
Martin Ester, Hans{-}Peter Kriegel, J{\"{o}}rg Sander, and Xiaowei Xu.
\newblock A density-based algorithm for discovering clusters in large spatial
  databases with noise.
\newblock In {\em Proceedings of the Second International Conference on
  Knowledge Discovery and Data Mining (KDD-96)}, pages 226--231, 1996.

\bibitem{Fiedler1998}
Miroslav Fiedler.
\newblock Ultrametric sets in euclidean point spaces.
\newblock {\em ELA. The Electronic Journal of Linear Algebra}, 3:23--30, 1998.

\bibitem{FischerB03}
Bernd Fischer and Joachim~M. Buhmann.
\newblock Path-based clustering for grouping of smooth curves and texture
  segmentation.
\newblock {\em IEEE Trans. Pattern Anal. Mach. Intell.}, 25(4):513--518, 2003.

\bibitem{doi:10.1198}
Chris Fraley and Adrian~E Raftery.
\newblock Model-based clustering, discriminant analysis, and density
  estimation.
\newblock {\em Journal of the American Statistical Association}, 97:611--631,
  2002.

\bibitem{2367502.2367564}
Lise Getoor and Ashwin Machanavajjhala.
\newblock Entity resolution: Theory, practice \& open challenges.
\newblock {\em Proc. VLDB Endow.}, 5(12), 2012.

\bibitem{Gower69}
C. Gower and G.~J.~S. Ross.
\newblock Minimum spanning trees and single linkage cluster analysis.
\newblock {\em Journal of the Royal Statistical Society. Series C}, 18(1),
  1969.

\bibitem{GuW10}
Yi Gu and Chaoli Wang.
\newblock A study of hierarchical correlation clustering for scientific volume
  data.
\newblock In {\em ISVC}, pages 437--446, 2010.

\bibitem{MortezaPhD}
{Haghir Chehreghani, Morteza}.
\newblock {\em Information-theoretic validation of clustering algorithms}.
\newblock PhD thesis, 2013.

\bibitem{jcgs2012}
Nicholas~A. Heard.
\newblock Iterative reclassification in agglomerative clustering.
\newblock {\em Journal of Computational and Graphical Statistics},
  20(4):920--936, 2012.

\bibitem{HoseiniRC21}
Fazeleh~Sadat Hoseini, Sadegh Rahrovani, and Morteza~Haghir Chehreghani.
\newblock Vehicle motion trajectories clustering via embedding transitive
  relations.
\newblock In {\em 24th {IEEE} International Intelligent Transportation Systems
  Conference, {ITSC}}, pages 1314--1321. {IEEE}, 2021.

\bibitem{Hu61}
T.C. Hu.
\newblock The maximum capacity route problem.
\newblock {\em Operations Research}, 9:898--900, 1961.

\bibitem{hubert1985comparing}
L. Hubert and P. Arabie.
\newblock {Comparing partitions}.
\newblock {\em Journal of classification}, 2(1):193--218, 1985.

\bibitem{HUGHES2004148}
Bruce Hughes.
\newblock Trees and ultrametric spaces: a categorical equivalence.
\newblock {\em Advances in Mathematics}, 189(1):148--191, 2004.

\bibitem{KalashnikovCMN08}
Dmitri~V. Kalashnikov, Zhaoqi Chen, Sharad Mehrotra, and Rabia Nuray{-}Turan.
\newblock Web people search via connection analysis.
\newblock {\em {IEEE} Trans. Knowl. Data Eng.}, 20(11):1550--1565, 2008.

\bibitem{CLUTO}
George Karypis.
\newblock Cluto - a clustering toolkit, 2002.
\newblock Retrieved from the University of Minnesota Digital Conservancy.

\bibitem{UCI_Datasets}
Markelle Kelly, Rachel Longjohn, and Kolby Nottingham.
\newblock The {UCI} machine learning repository.
\newblock \url{http://archive.ics.uci.edu/datasets}.

\bibitem{KimC07icml}
Kye{-}Hyeon Kim and Seungjin Choi.
\newblock Neighbor search with global geometry: a minimax message passing
  algorithm.
\newblock In {\em Twenty-Fourth International Conference on Machine Learning,
  ICML}, pages 401--408, 2007.

\bibitem{KimNKY11}
Sungwoong Kim, Sebastian Nowozin, Pushmeet Kohli, and Chang~Dong Yoo.
\newblock Higher-order correlation clustering for image segmentation.
\newblock In {\em Advances in Neural Information Processing Systems 24 (NIPS)},
  pages 1530--1538, 2011.

\bibitem{KimYNK14}
Sungwoong Kim, Chang~Dong Yoo, Sebastian Nowozin, and Pushmeet Kohli.
\newblock Image segmentation using higher-order correlation clustering.
\newblock {\em {IEEE} Trans. Pattern Anal. Mach. Intell.}, 36(9):1761--1774,
  2014.

\bibitem{biom12647}
Patrick~K. Kimes, Yufeng Liu, David Neil~Hayes, and James~Stephen Marron.
\newblock Statistical significance for hierarchical clustering.
\newblock {\em Biometrics}, 73(3):811--821, 2017.

\bibitem{lance67hierarchical}
G.~N. Lance and W.~T. Williams.
\newblock A general theory of classificatory sorting strategies.
\newblock {\em The Computer Journal}, 9(4):373--380, 1967.

\bibitem{Leclerc1981}
Bruno Leclerc.
\newblock Description combinatoire des ultramétriques.
\newblock {\em Mathématiques et Sciences Humaines}, 73:5--37, 1981.

\bibitem{Levenstien2003}
Mark~A. Levenstien, Yaning Yang, and Jurg Ott.
\newblock Statistical significance for hierarchical clustering in genetic
  association and microarray expression studies.
\newblock {\em BMC Bioinformatics}, 4(1), 2003.

\bibitem{LiebmannWS18}
Tom Liebmann, Gunther~H. Weber, and Gerik Scheuermann.
\newblock Hierarchical correlation clustering in multiple 2d scalar fields.
\newblock {\em Comput. Graph. Forum}, 37(3):1--12, 2018.

\bibitem{LittleMM20}
Anna~V. Little, Mauro Maggioni, and James~M. Murphy.
\newblock Path-based spectral clustering: Guarantees, robustness to outliers,
  and fast algorithms.
\newblock {\em J. Mach. Learn. Res.}, 21:6:1--6:66, 2020.

\bibitem{mcqueen1967smc}
J. MacQueen.
\newblock Some methods for classification and analysis of multivariate
  observations.
\newblock In {\em 5th Berkeley Symposium on Mathematical Statistics and
  Probability}, pages 281--297, 1967.

\bibitem{Maimon:2005}
Oded Maimon and Lior Rokach.
\newblock {\em Data Mining and Knowledge Discovery Handbook}.
\newblock 2005.

\bibitem{MartnezPrez2008}
{\'{A}}lvaro Mart{\'{\i}}nez-P{\'{e}}rez and Manuel~A. Mor{\'{o}}n.
\newblock Uniformly continuous maps between ends of
  {\textdollar}{\textdollar}$\lbrace${\textbackslash}mathbb$\lbrace$r$\rbrace$$\rbrace${\textdollar}{\textdollar}
  -trees.
\newblock {\em Mathematische Zeitschrift}, 263(3):583--606, 2008.

\bibitem{MS10}
Claire Mathieu and Warren Schudy.
\newblock Correlation clustering with noisy input.
\newblock In {\em SODA '10}, pages 712--728, 2010.

\bibitem{MDS}
A. Mead.
\newblock Review of the development of multidimensional scaling methods.
\newblock {\em Journal of the Royal Statistical Society: Series D},
  41(1):27--39, 1992.

\bibitem{abs-1109-2378}
Daniel Mullner.
\newblock Modern hierarchical, agglomerative clustering algorithms.
\newblock {\em CoRR}, abs/1109.2378, 2011.

\bibitem{Nadler07}
Boaz Nadler and Meirav Galun.
\newblock Fundamental limitations of spectral clustering.
\newblock In {\em Advances in Neural Information Processing Systems (NIPS)},
  pages 1017--1024, 2007.

\bibitem{Ng01onspectral}
Andrew~Y. Ng, Michael~I. Jordan, and Yair Weiss.
\newblock On spectral clustering: Analysis and an algorithm.
\newblock In {\em Advances in Neural Information Processing Systems (NIPS)},
  pages 849--856, 2001.

\bibitem{Peng2022}
Dehua Peng, Zhipeng Gui, Dehe Wang, Yuncheng Ma, Zichen Huang, Yu Zhou, and
  Huayi Wu.
\newblock Clustering by measuring local direction centrality for data with
  heterogeneous density and weak connectivity.
\newblock {\em Nature Communications}, 13(1), 2022.

\bibitem{podani2000introduction}
J. Podani.
\newblock {\em Introduction to the exploration of multivariate biological
  data}.
\newblock Backhuys Publishers, 2000.

\bibitem{Roy:2017:HCV}
Aurko Roy and Sebastian Pokutta.
\newblock Hierarchical clustering via spreading metrics.
\newblock {\em J. Mach. Learn. Res.}, 18(1):3077--3111, 2017.

\bibitem{Schoenberg}
I.~J. Schoenberg.
\newblock On certain metric spaces arising from euclidean spaces by a change of
  metric and their imbedding in hilbert space.
\newblock {\em Annals of Mathematics}, 38(4):787--793, 1937.

\bibitem{ShiDELM21}
Jessica Shi, Laxman Dhulipala, David Eisenstat, Jakub Lacki, and Vahab~S.
  Mirrokni.
\newblock Scalable community detection via parallel correlation clustering.
\newblock {\em Proc. {VLDB} Endow.}, 14(11):2305--2313, 2021.

\bibitem{Shi:2000:NCI}
Jianbo Shi and Jitendra Malik.
\newblock Normalized cuts and image segmentation.
\newblock {\em IEEE Trans. Pattern Anal. Mach. Intell.}, 22(8):888--905, 2000.

\bibitem{ShiehHA11}
A. Shieh, T.~B. Hashimoto, and E.~M. Airoldi.
\newblock Tree preserving embedding.
\newblock In {\em International Conference on Machine Learning (ICML)}, pages
  753--760, 2011.

\bibitem{Shieh16916}
A.~D. Shieh, T.~B. Hashimoto, and E.~M. Airoldi.
\newblock Tree preserving embedding.
\newblock {\em Proceedings of the National Academy of Sciences},
  108(41):16916--16921, 2011.

\bibitem{sneath1957dn09j}
Peter Henry~Andrews Sneath.
\newblock The application of computers to taxonomy.
\newblock {\em Journal of General Microbiology}, 17:201--226, 1957.

\bibitem{sokal58}
R.~R. Sokal and C.~D. Michener.
\newblock A statistical method for evaluating systematic relationships.
\newblock {\em Uni. of Kansas Science Bulletin}, 38:1409--1438, 1958.

\bibitem{TangCAL16}
Jiliang Tang, Yi Chang, Charu~C. Aggarwal, and Huan Liu.
\newblock A survey of signed network mining in social media.
\newblock {\em {ACM} Comput. Surv.}, 49(3):42:1--42:37, 2016.

\bibitem{ThielCD19}
Erik Thiel, Morteza~Haghir Chehreghani, and Devdatt~P. Dubhashi.
\newblock A non-convex optimization approach to correlation clustering.
\newblock In {\em The Thirty-Third {AAAI} Conference on Artificial Intelligence
  (AAAI)}, pages 5159--5166, 2019.

\bibitem{Titterington1985-uj}
D.~M. Titterington, {etc.}, A.~F.~M. Smith, and U.~E. Makov.
\newblock {\em Statistical analysis of finite mixture distribution}.
\newblock Probability \& Mathematical Statistics S. John Wiley \& Sons, 1985.

\bibitem{VainsteinCCRMA21}
D. Vainstein, V. Chatziafratis, G. Citovsky, A. Rajagopalan, M. Mahdian, and Y.
  Azar.
\newblock Hierarchical clustering via sketches and hierarchical correlation
  clustering.
\newblock In {\em {AISTATS}}, pages 559--567, 2021.

\bibitem{vandermaaten08a}
Laurens van~der Maaten and Geoffrey Hinton.
\newblock Visualizing data using t-sne.
\newblock {\em Journal of Machine Learning Research}, 9(86):2579--2605, 2008.

\bibitem{Varga1993OnSU}
Richard~S. Varga and Reinhard Nabben.
\newblock On symmetric ultrametric matrices.
\newblock In {\em Numerical Linear Algebra}, 1993.

\bibitem{Vinh:2010}
Nguyen~Xuan Vinh, Julien Epps, and James Bailey.
\newblock Information theoretic measures for clusterings comparison: Variants,
  properties, normalization and correction for chance.
\newblock {\em J. Mach. Learn. Res.}, 11:2837--2854, 2010.

\bibitem{Luxburg07}
Ulrike von Luxburg.
\newblock A tutorial on spectral clustering.
\newblock {\em Statistics and Computing}, 17(4):395--416, 2007.

\bibitem{abs-1708-07747}
Han Xiao, Kashif Rasul, and Roland Vollgraf.
\newblock Fashion-mnist: a novel image dataset for benchmarking machine
  learning algorithms.
\newblock {\em CoRR}, abs/1708.07747, 2017.

\bibitem{Yildirim2017}
Pelin Yildirim and Derya Birant.
\newblock K-linkage: A new agglomerative approach for hierarchical clustering.
\newblock 17:77--88, 2017.

\bibitem{RePEc1938}
Gale Young and A. Householder.
\newblock Discussion of a set of points in terms of their mutual distances.
\newblock {\em Psychometrika}, 3(1):19--22, 1938.

\end{thebibliography}
}
\newpage
\hspace{1mm}
\newpage
\appendix

\section{\label{Appendix:HCC_Alg} HCC Algorithm}

Algorithm \ref{alg:hierarchical_CC} describes hierarchical correlation clustering (HCC) in detail.
The algorithm at the beginning assumes $n$ singleton clusters, one for each object. For each cluster, it obtains the nearest cluster and the respective similarity. The algorithm then iteratively performs the following steps. i) Finds the two nearest clusters according to the inter-cluster (dis)similarity function defined in Eq. \ref{eq:distCC}. ii) Merges the respective clusters to build a new cluster at a higher level.  iii) Updates the inter-cluster similarity matrix $\mathbf S$, the nearest neighbor vector $nn\_ind$ and the respective similarities $nn\_sim$.\footnote{In our implementation, we use a data structure similar to the linkage matrix used by \emph{scipy} package in Python to encode the dendrogram and store the intermediate clusters.}

\begin{algorithm}[thb!]
\caption{Hierarchical Correlation Clustering.}
\label{alg:hierarchical_CC}
\begin{algorithmic} [1]
\REQUIRE {A set of $n$ objects $\mathbf O = \{0,..., n-1\}$ and the pairwise similarities $\mathbf S$.}

\FORALL{$i \in \mathbf O$}
\STATE $nn\_ind[i] = \arg\max_j \mathbf S[i,j]$
\STATE $nn\_sim[i]= \max_j \mathbf S[i,j]$
\ENDFOR

\STATE $n\_c = |\mathbf O|$   \COMMENT{shows the number of active clusters}

\WHILE{$n\_c > 1$}

\STATE\COMMENT{find the indices of the two nearest clusters, w.l.g. we assume $u \le v$}
\STATE $u = \arg\max_i nn\_sim[i]$
\STATE $v = nn\_ind[u]$

\STATE\COMMENT{update the inter-cluster (dis)similarities, active clusters and other parameters}
\FORALL{$i \in \{0, ..., n\_c  \}  $}
\STATE $new\_sim[i] = \mathbf S[i,u] +  \mathbf S[i,v]$
\ENDFOR

\STATE Remove($new\_sim[v]$)
\STATE Remove($new\_sim[u]$)

\STATE Remove($\mathbf S[v,:]$)
\STATE Remove($\mathbf S[:,v]$)

\STATE Remove($\mathbf S[u,:]$)
\STATE Remove($\mathbf S[:,v]$)

\STATE Append($\mathbf S, new\_sim $)
\STATE Append($\mathbf S, [new\_sim, 0]^T$)

\STATE $n\_c = n\_c -1$

\STATE Remove($nn\_ind[v]$)
\STATE Remove($nn\_sim[u]$)

\STATE Update ($nn\_ind$)
\STATE Update ($nn\_sim$)
\STATE Append($nn\_ind, \arg\max_j \mathbf S[n\_c,j]$)
\STATE Append($nn\_sim, \max_j \mathbf S[n\_c,j]$)

\ENDWHILE

Return the intermediate clusters and the dendrogram.
\end{algorithmic}
\end{algorithm}

\section{\label{Appendix:Proofs}Proofs}

\subsection{Proof of Lemma 1}
One can show that the pairwise minimax dissimilarities across any given graph are identical to the pairwise minimax dissimilarities present in any minimum spanning tree obtained from the same graph. The proof is similar to the \emph{maximum capacity} problem \cite{Hu61}. Thereby, the minimax dissimilarities are obtained by

\begin{eqnarray}
	\mathbf D_{i,j}^{MM} &=& \min_{p\in \mathcal P_{ij}(\mathcal G)}\left\{ \max_{1\le l \le |p|-1}\mathbf D_{p(l)p(l+1)}\right\} \nonumber \\
	&=&  \max_{1\le l \le |p_{ij}|-1}\mathbf D_{p(l)p(l+1)},
\label{Eq.pathTree}
\end{eqnarray}

where $p_{ij}$ indicates the (only) path between $i$ and $j$ on a minimum spanning tree computed on $\mathcal G$. To obtain the minimax dissimilarities $\mathbf D^{MM}_{ij}$, we can just select the maximal edge weight on the only path between $i$ and $j$ on the minimum spanning tree.

On the other hand, the single linkage method and the Kruskal's minimum spanning tree algorithm are equivalent \cite{Gower69}.
Thus, the dendrogram $T$ obtained via single linkage sufficiently contains the pairwise minimax dissimilarities. Now, we elaborate that the minimax dissimilarities in Eq. \ref{Eq.pathTree} equal the dissimilarities defined in Eq. \ref{eq:linkage_hierarchy_dist}, i.e.,  $\mathbf X_{ij}$ is the largest edge weight on the path between $i$ and $j$ in the hierarchy.

Given $i,j$, let
\begin{equation}
T^* = \arg\min linkage(T') \quad \text{s.t. }  i,j\in T'  \text{ and } T'\in \mathcal{T}_T \, .
\end{equation}
Then, $T^*$ represents a minimum spanning subtree, which includes a path between $i$ and $j$ (because the root cluster of $T^*$ contains both $i$ and $j$) and it is consistent with a complete minimum spanning on all the objects. On the other hand, we know that for each pair of clusters $\mathbf u,  \mathbf v\ \in T^*$ which have direct or indirect parent-child relation, we have, $linkage(\mathbf u) \ge linkage(\mathbf v)$ iff $level(\mathbf u) \ge level(\mathbf v)$. This implies  the linkage of the root cluster of $T^*$ represents the maximal edge weight on the path between $i$ and $j$ represented by the dendrogram $T$. Thus,  $\mathbf X_{ij}$  represents $\mathbf D^{MM}_{ij}$

Thereby, minimax dissimilarities correspond to building a single linkage dendrogram and using the linkage as the pairwise dissimilarity between the objects in the two respective clusters. We know that according to Proposition \ref{thm:agg_shift} single linkage dendrogram is shift-invariant. Therefore, by shifting the pairwise dissimilarities by a sufficient $\alpha$, there will be no change in the paths between the clusters of single linkage dendrogram, nor in the paths representing the minimax dissimilarities.

\subsection{Proof of Theorem 2}
Over a graph, we define a path between $i$ and $j$ to be \emph{positive} if all the edge weights on the path are positive. Then, we have the following observations.

\begin{enumerate}[leftmargin=*]
  \item On a general graph $\mathcal G(\mathbf O, \mathbf S)$, one can see that in the optimal solution of correlation clustering, if the two objects $i$ and $j$ are in the same cluster, then there is at least one positive path between them (the proof can be done by contradiction; if there is no such a path, then the two objects should be in separate clusters in order to avoid the increase in the cost function).
  \item Whenever there is a positive path between $i$ and $j$, then their minimax similarity $\mathbf S^{MM}_{i,j}$ will be necessarily positive. Therefore, when we apply correlation clustering to graph $\mathcal G(\mathbf O, \mathbf S^{MM})$, all the intra-cluster similarities of the optimal clusters will be positive. This corresponds to having a positive path between every two objects that are in the same optimal cluster, i.e., they are in the same connected component of $\mathcal G(\mathbf O, \mathbf S')$ where $\mathbf S'$ is defined as 

\begin{equation}
  \mathbf S'_{ij}=\begin{cases}
    1, & \text{if $\mathbf S_{ij} >0$}.\\
   0, & \text{otherwise}.
  \end{cases}
  \label{eq:connectedCMatrix}
\end{equation}
  
  \item We can also conclude that when we apply correlation clustering to graph $\mathcal G(\mathbf O, \mathbf S^{MM})$, then for any optimal cluster $\mathbf c$, there is no object $i \notin \mathbf c$ such that $i$ has a positive path to an object in $\mathbf c$. Otherwise, $i$ and all the other objects outside $\mathbf c$ with positive paths to $i$ would have positive paths to all the objects in  $\mathbf c$ such that all of them should be clustered together.
\end{enumerate}

Now we study the connection of connected components of graph $\mathcal G(\mathbf O, \mathbf S')$ to the optimal correlation clustering on $\mathcal G(\mathbf O, \mathbf S^{MM})$.

\begin{itemize}[leftmargin=*]
  \item There is a positive path between every two objects in a connected component of $\mathcal G(\mathbf O, \mathbf S')$. Thus, they are in the same optimal cluster of $\mathcal G(\mathbf O, \mathbf S^{MM})$.
  \item If two nodes $i$ and $j$ are at two different connected components, then there is no positive path between them either on $\mathcal G(\mathbf O, \mathbf S')$  or on $\mathcal G(\mathbf O, \mathbf S^{MM})$. Thus, they cannot be in the same cluster if we apply correlation clustering on $\mathcal G(\mathbf O, \mathbf S^{MM})$.
\end{itemize}

Thus, we conclude that the connected components of  $\mathcal G(\mathbf O, \mathbf S')$ correspond to the optimal correlation clustering on graph $\mathcal G(\mathbf O, \mathbf S^{MM})$.

\begin{figure*}[htb!]
    \centering
    \begin{subfigure}{0.30\linewidth}
        \includegraphics[width=1\textwidth]{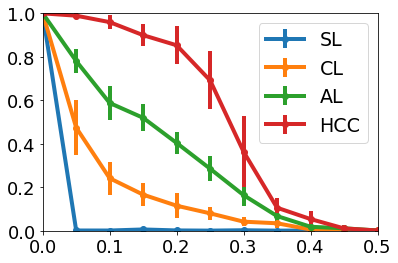}
        \caption{\emph{Breast Tissue}}
        \label{fig:Breast_Tissue_Rand}
    \end{subfigure}
    \begin{subfigure}{0.30\linewidth}
        \includegraphics[width=1\textwidth]{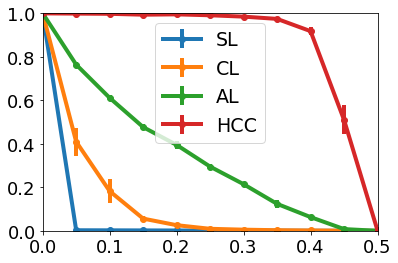}
        \caption{\emph{Cardiotocography}}
        \label{fig:Cardiotocography_Rand}
    \end{subfigure}
    \begin{subfigure}{0.30\linewidth}
        \includegraphics[width=1\textwidth]{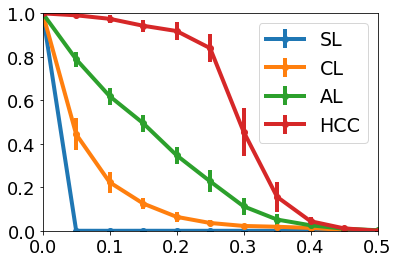}
        \caption{\emph{Image Segmentation}}
        \label{fig:Image_Segmentation_Rand}
    \end{subfigure}
    \\
    \begin{subfigure}{0.30\linewidth}
        \includegraphics[width=1\textwidth]{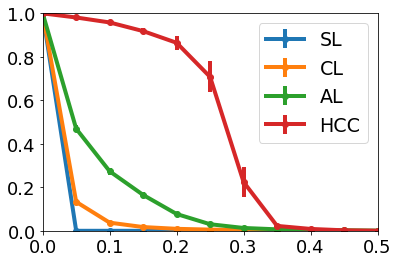}
        \caption{\emph{ISOLET}}
        \label{fig:ISOLET5_Rand}
    \end{subfigure}
    \begin{subfigure}{0.30\linewidth}
        \includegraphics[width=1\textwidth]{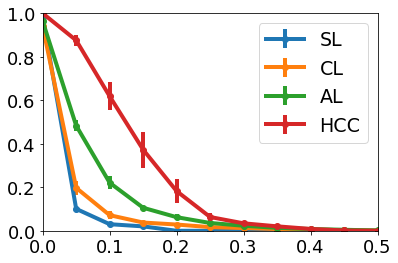}
        \caption{\emph{Leaf}}
        \label{fig:Leaf_Rand}
    \end{subfigure}
    \begin{subfigure}{0.30\linewidth}
        \includegraphics[width=1\textwidth]{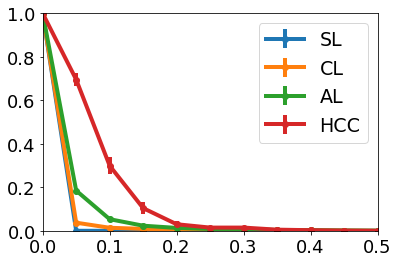}
        \caption{\emph{One-Hundred Plant}}
        \label{fig:One_hundred_plant_Rand}
    \end{subfigure}
    \caption{Rand score of  hierarchical clustering methods applied to the UCI datasets (the x-axis shows the flip noise parameter $\eta$).
    Similar to the MI measure, HCC  provides the best scores, even when the datasets are difficult to cluster.
    }
    \label{fig:UCI_hierarchical_Rand}
\end{figure*}

\subsection{Proof of Theorem 3}

The  approximate algorithm in \cite{AilonCN08} iteratively picks an unclustered object and its positive neighbors as a new cluster. According to Theorem \ref{thm:CC_MM}, the optimal solution of correlation clustering applied to $\mathcal G(\mathbf O, \mathbf S^{MM})$ corresponds to extracting the connected components of graph $\mathcal G(\mathbf O, \mathbf S')$, where $\mathbf S'$ is defined in Eq. \ref{eq:connectedCMatrix}.

Thus it is sufficient to show the two followings.  
\begin{enumerate}
    \item If the algorithm  in \cite{AilonCN08} on $\mathcal G(\mathbf O, \mathbf S^{MM})$  picks  $i$ and $j$  in the same cluster then  $\mathbf S^{MM}_{i,j} = +1$. This indicates that $i$ and $j$ have a positive path on $\mathcal G(\mathbf O, \mathbf S)$ (a positive path is defined in Theorem \ref{thm:CC_MM}), i.e., $i$ and $j$ are in the same connected component of  $\mathcal G(\mathbf O, \mathbf S')$.
    \item If $i$ and $j$ are in different clusters according to algorithm  \cite{AilonCN08} applied to $\mathcal G(\mathbf O, \mathbf S^{MM})$, then $\mathbf S^{MM}_{i,j} = -1$. This indicates that there is no  positive path between $i$ and $j$ on $\mathcal G(\mathbf O, \mathbf S)$  and also on $\mathcal G(\mathbf O, \mathbf S^{MM})$, i.e., $i$ and $j$ are in different connected component of  $\mathcal G(\mathbf O, \mathbf S')$.
\end{enumerate}

\section{\label{Appendix:AddExperiments}Additional Experimental Results}
\subsection{More results with UCI datasets}
In Figure \ref{fig:UCI_hierarchical_Rand}, we demonstrate the performance of HCC on different UCI datasets w.r.t. Rand score, and compare the results with other agglomerative methods (the results w.r.t. MI are shown in the main text in Figure \ref{fig:UCI_hierarchical_MI}). We observe that consistent with the MI measure, HCC yields the best scores, even at high noise levels and when the datasets are difficult to cluster.

\begin{table*}[htb]
\caption{Performance of different hierarchical clustering methods on 20 newsgroup datasets.
HCC yields the best results according to the different evaluation measures.
}
\centering 
\begin{tabular}{|| c || c c || c c || c c || c c ||} 
\hline\hline 
 &\multicolumn{2}{c||}{news1}&\multicolumn{2}{c||}{news2}&\multicolumn{2}{c||}{news3}&\multicolumn{2}{c||}{news4}\\
 \hline
 method & MI & Rand &MI & Rand & MI & Rand & MI & Rand \\ 
\hline 
SL & 0.034 & 0.017 & 0.043 & 0.039 & 0.021 & 0.044 & 0.020 & 0.038  \\
CL & 0.266 & 0.277 & 0.255 & 0.230 & 0.501 & 0.594 & 0.121 & 0.116  \\
AL & 0.287 & 0.228 & 0.342 & 0.344 & 0.685 & 0.750 & 0.498 & \bf{0.548} \\
HCC & \bf{0.331} & \bf{0.287} & \bf{0.370} & \bf{0.368} & \bf{0.794} & \bf{0.854} & \bf{0.541} & 0.499 \\
\hline 
\end{tabular}
\label{table:newsgroup_hierarchy} 
\end{table*}

\begin{table*}[htb!]
\caption{Performance of tree-preserving embedding methods on different 20 newsgroup datasets applied with GMM.
The embeddings obtained by HCC yield better results.
}
\centering 
\begin{tabular}{|| c || c c || c c || c c || c c ||} 
\hline\hline 
 &\multicolumn{2}{c||}{news1}&\multicolumn{2}{c||}{news2}&\multicolumn{2}{c||}{news3}&\multicolumn{2}{c||}{news4}\\
 \hline
 method & MI & Rand & MI & Rand & MI & Rand & MI & Rand \\ 
\hline 
SL               & 0.034 & 0.017 & 0.043 & 0.039 & 0.021 & 0.044 & 0.020 & 0.038  \\
SL+GMM & 0.191 & 0.158 & 0.108 & 0.097 & 0.166 & 0.210 & 0.134 & 0.120 \\
\hline
CL & 0.266 & 0.277 & 0.255 & 0.230 & 0.501 & 0.594 & 0.121 & 0.116    \\
CL+GMM & 0.271 & 0.275 & 0.272 & 0.239 & 0.522 & 0.587 & 0.118 & 0.119 \\
\hline
AL & 0.287 & 0.228 & 0.342 & 0.344 & 0.685 & 0.750 & 0.498 & \bf{0.548} \\
AL+GMM & 0.309 & 0.279 & 0.358 & 0.350 & 0.701 & 0.773 & 0.503 & 0.525 \\
\hline
HCC & 0.331 & 0.287 & 0.370 & 0.368 & 0.794 & 0.854 & 0.541 & 0.499 \\
HCC+GMM & \bf{0.344} & \bf{0.297} & \bf{0.439} & \bf{0.443} & \bf{0.831} & \bf{0.892} & \bf{0.560} & 	{0.519} \\
\hline 
\end{tabular}
\label{table:newsgroup_embedding_GMM} 
\end{table*}

\begin{table}[hbt]
\caption{Performance of different methods on webpage dataset. The embeddings obtained by HCC yield better results.
}
\centering 
\begin{tabular}{|| c || c c ||} 
\hline\hline 
 method & MI & Rand \\
\hline 
SL               & 0.218 & 0.192 \\
SL+GMM & 0.254 & 0.211  \\
\hline
CL               & 0.386 & 0.417 \\
CL+GMM & 0.392 & 0.405 \\
\hline
AL              & 0.459 & 0.488 \\
AL+GMM & 0.472 & 0.491 \\
\hline
HCC               & 0.583 & 0.575 \\
HCC+GMM & \bf{0.622} & \bf{0.630} \\
\hline 
\end{tabular}
\label{table:web_pages} 
\end{table}

\subsection{HCC on 20 newsgroup data}

In the following, we study the performance of different methods on several subsets of 20 newsgroup data collection chosen randomly from different categories.

\begin{enumerate}
\item \emph{news1}: the $3901$ documents of the categories 'misc.forsale', 'rec.motorcycles', 'talk.politics.mideast', 'sci.med' ($48596$ dimensions).
\item \emph{news2}: the $3743$ documents of the categories 'alt.atheism', 'comp.sys.mac.hardware', 'sci.electronics', 'soc.religion.christian' ($40735$ dimensions).
\item \emph{news3}: the $1984$ documents of the categories 'sci.space', 'soc.religion.christian'  ($30749$ dimensions).
\item \emph{news4}: the $2877$ documents of the categories  'comp.graphics', 'rec.sport.baseball', 'talk.politics.guns' ($38177$ dimensions).
\end{enumerate}

For each dataset, we compute the TF-IDF vectors of the documents and apply PCA with $50$ principal components.  We obtain the similarity between every two documents via the cosine similarity between their respective PCA vectors. As has been discussed in detail in \cite{MortezaPhD}, adding a fixed number to all the pairwise similarities can possibly improve the clustering results.\footnote{It is suggested in \cite{MortezaShift_MLj}  to adaptively shift the pairwise similarities so that the sum of pairwise similarities equals zero for each object in the dataset.} 
Table \ref{table:newsgroup_hierarchy} shows the results for various hierarchical clustering methods w.r.t. different  evaluation criteria. Among the different methods, HCC usually yields the best results, and AL is the second best choice.

In the following, we investigate tree-preserving embedding on these datasets. Table \ref{table:newsgroup_embedding_GMM} presents the results of tree-preserving embedding compared to hierarchical clustering. Specifically, we investigate the benefits of using hierarchical clustering for feature extraction. We observe, i) employing hierarchical clustering to extract features for a method such as GMM usually yields improving the results, and ii) HCC, whether used directly for clustering or for feature extraction, often gives superior results compared to the alternatives. 

\subsection{Experiments with web data}
Finally, we investigate HCC for both hierarchical clustering and feature extraction on a web dataset. The dataset consists of $15,000$ webpages collected about topics such as politics, finance, sport, art, entertainment, health, technology, environment, cars and films. Similar to 20 newsgroup datasets, we compute the TF-IDF vectors,  apply PCA and then obtain the pairwise cosine similarities between the web pages. Table \ref{table:web_pages} demonstrates the performance of different methods on this dataset. We observe that, consistent with the previous experiments, both HCC and HCC+GMM yield improving the results compared to the baselines. In addition, using HCC to compute intermediate features for GMM (i.e., HCC+GMM) results in better scores than using HCC to produce the final clusters.
\end{document}